\documentclass{article}

\usepackage[preprint,nonatbib]{neurips_2020}

\usepackage[utf8]{inputenc} 
\usepackage[T1]{fontenc}    
\usepackage{url}            
\usepackage{booktabs}       
\usepackage{amsfonts}       
\usepackage{nicefrac}       
\usepackage{microtype}      
\usepackage{subfigure}

\usepackage{xcolor}
\usepackage{soul}
\usepackage{graphicx}
\usepackage[linesnumbered, ruled]{algorithm2e}
\usepackage{multirow}
\usepackage{makecell}
\usepackage{enumitem}
\usepackage{amsmath}
\usepackage{float}
\usepackage{bm}

\usepackage{array}
\newcolumntype{L}[1]{>{\raggedright\let\newline\\\arraybackslash\hspace{0pt}}m{#1}}
\newcolumntype{C}[1]{>{\centering\let\newline  \\\arraybackslash\hspace{0pt}}m{#1}}
\newcolumntype{R}[1]{>{\raggedleft\let\newline \\\arraybackslash\hspace{0pt}}m{#1}}

\title{Simplify and Robustify Negative Sampling for Implicit Collaborative Filtering}

\author{%
	Jingtao Ding$^1$\thanks{The first three authors have equal contributions.}
	\quad
	Yuhan Quan$^1$
	\quad
	Quanming Yao$^2$
	\quad
	Yong Li$^1$
	\quad
	Depeng Jin$^1$\\
	$^1$Tsinghua University,
	\quad
	$^2$4Paradigm Inc (Hong Kong)
}

\begin{document}

\maketitle

\begin{abstract}
Negative sampling approaches are prevalent in implicit collaborative filtering for obtaining negative labels from massive unlabeled data.
As two major concerns in negative sampling, efficiency and effectiveness are still not fully achieved by recent works that use complicate structures and overlook risk of false negative instances.
In this paper, we first provide a novel understanding of negative instances by empirically observing that only a few instances 
are potentially important for model learning, and false negatives tend to have stable predictions over many training iterations.
Above findings motivate us to simplify the model by sampling from designed memory that only stores a few important candidates and, more importantly, 
tackle the untouched false negative problem by favouring high-variance samples stored in memory, which achieves efficient sampling of true negatives with high-quality.
Empirical results on two synthetic datasets and three real-world datasets demonstrate both robustness and superiorities of our negative sampling method.

\end{abstract}

\section{Introduction}
Collaborative filtering~(CF), 
as the key technique of personalized recommender systems, 
focuses on learning user preference from the observed user-item interactions~\cite{OCCF,BPR}. 
Today's recommender systems also witness the prevalence of implicit user feedback, 
such as purchases in E-commerce sites and watches in online video platforms, 
which is much easier to collect compared to the explicit feedback~(such as ratings) on item utility.
In above examples, each observed interaction normally indicates a user's interest on an item, 
\textit{i.e.}, a positive label, while the rest unobserved interactions are unlabeled.
As for learning an implicit CF model from this positive-only data, 
a widely adopted approach is to select a few instances from the unlabeled
part and treat them as negative labels, 
also known as negative sampling~\cite{NNCF,BPR}. 
Then,
the CF model is optimized to give positive instances higher scores than those given to negative ones~\cite{BPR}.

Similar to other related applications in representation learning of text~\cite{word2vec} or graph data~\cite{deepwalk}, 
negative sampling in implicit CF also has two major concerns, 
i.e., 
efficiency and effectiveness~\cite{NNCF,DNS}.
First, the efficient sampling process is required, 
as the number of unobserved user-item interactions can be extremely huge. 
Second, the sampled instances need to be high-quality, so as to learn useful information about user's negative preference.
However, since implicit CF is an application-driven problem where user behaviors play an important role, it may be unrealistic to assume that unobserved interactions are all negative, which introduces false negative instances into training process~\cite{hernandez2014probabilistic,ExpoMF,zhao2018interpreting}. 
For example, 
an item may be ignored because of its displayed position and form, not necessarily the user's dislike.
Therefore, 
false negative instances naturally exist in implicit CF.

Previous works of negative sampling in implicit CF mainly focus on replacing the uniform sampling distribution with another proposed distribution, so as to improve the quality of negative samples.   
Similar to the word-frequency based distribution~\cite{word2vec} and node-degree based distribution~\cite{deepwalk} used in other domains, an item-popularity based distribution that favours popular items is usually adopted~\cite{NNCF,NCE4CF}.
In terms of sample quality, the strategy emphasizing hard negative samples has been proven to be more effective~\cite{AdvIR}, as it can bring more information for model training.
Specifically, this is achieved by either assigning higher probability to 
instances with large prediction score~\cite{AdaOverSam,DNS} 
or leveraging techniques of adversarial learning~\cite{RNS,AdvIR,IRGAN}.
Nevertheless, 
the above hard negative sampling approaches cannot simultaneously meet the requirements on  
efficiency and effectiveness.
On the one hand, 
several state-of-the-art solutions~\cite{RNS,AdvIR} use complicate structures like generative adversarial network~(GAN)~\cite{GAN}
for generating negative instances, 
which has posed a severe challenge on model efficiency.  
On the other hand, 
all these methods overlook the risk of introducing false negative instances and instead only focus on hard ones, 
making the sampling process less robust for training an effective CF model with false negatives.

Different from above works, this paper formulates the negative sampling problem as efficient learning from unlabeled data with the presence of noisy labels, \textit{i.e.}, false negative instances.
We propose to simplify and robustify the negative sampling for implicit CF, 
which has three main challenges:
\begin{itemize}[leftmargin=*,noitemsep,topsep=0pt,parsep=0pt,partopsep=0pt]
\item 
\textbf{How to capture the dynamic distribution of true negative instances with a simple model?} 
In the implicit CF problem, true negative instances are hidden inside the massive unlabeled data, along with false negative instances.
Although negative instances in other domains follow a skewed distribution and can be modeled by a simple model~\cite{word2vec,NSCaching}, it remains unknown if this prior knowledge can be applied in the implicit CF problem that expects true negative instances only.

\item \textbf{ How can we reliably measure the quality of negative samples? } 
Given the risk of introducing false negative instances, the quality of negative samples needs to be measured in a more reliable way.
However, it is non-trivial to design a discriminative criterion that can help to accurately identify true negative instances with high quality.

\item \textbf{How can we efficiently sample true negative instances of high-quality?} 
Although learning effective information from unlabeled and noisy data is related to general machine learning approaches including positive-unlabeled leaning~\cite{nnPU} and instance re-weighting~\cite{L2Reweight}, 
these methods are not suitable for implicit CF problem, where the huge number of unobserved user-item interactions requires an efficient modeling.
Instead, our proposed method needs to maintain both efficiency, by sampling, and effectiveness, by considering samples' informativeness and reliability simultaneously.
This has not been tackled before in both implicit CF and other similar problems.
\end{itemize}

Solving above three challenges calls for a deep and fundamental understanding of different negative instances in implicit CF problem.
In this paper, we empirically find that negative instances with large prediction scores are important for the model learning but generally rare, \textit{i.e.}, following a skewed distribution.
A more novel finding is that false negative instances always have large scores over many iterations of training, \textit{i.e.}, a lower variance, which provides a new angle on tackling false negative problem remained in existing approaches.
Motivated by above two findings, we propose a novel simplified and robust negative sampling approach, named SRNS, that
1) captures the dynamic distribution of negative instances with a memory-based model, by simply maintaining the promising candidates with large scores, and
2) leverages a high-variance based criterion to reliably measure the quality of negative samples, reducing the risk of false negative instances effectively.
Above two designs are further combined into a two-step sampling scheme that constantly alternates between score-based memory update and variance-based sampling, so as to efficiently sample true negative instances with high-quality.
Experiment results on two synthetic datasets demonstrate the robustness of our SRNS under various levels of noisy circumstances.
Further experiments on three real-world datasets also empirically validates its superiorities over state-of-the-art baselines, in terms of effectiveness and efficiency.

\section{Background}
\label{sec:negsap}

Training an implicit CF model generally involves three main steps, \textit{i.e.}, choosing scoring function $r$, objective function $L$ and negative sampling distribution $p_\mathrm{ns}$.
The scoring function $r(\textbf{p}_u, \textbf{q}_i, \bm{\beta})$ calculates the relevance between a user $u\in\mathcal{U}$ and an item $i\in\mathcal{I}$ based on $u$'s embedding $\mathbf{p}_u \in \mathbb{R}^F$ and $i$'s embedding $\mathbf{q}_{i} \in \mathbb{R}^F$, with a learnable parameter $\bm{\beta}$.
It can be chosen among various candidates including matrix factorization~(MF)~\cite{koren2008factorization}, 
multi-layer perceptron~(MLP)~\cite{NCF}, 
graph neural network~(GNN)~\cite{berg2017graph,wang2019neural}, etc.
For example, the generalized matrix factorization~(GMF)~\cite{NCF} is: 
$r(\textbf{p}_u, \textbf{q}_i, \bm{\beta}) = {\bm{\beta}}^{\top}  (\textbf{p}_u \odot \textbf{q}_i)$,
where the learnable parameter of $r$ is a vector ${\bm{\beta}}$ and $\odot $ denotes element-wise product.
A large value of $r(\textbf{p}_u, \textbf{q}_i, \bm{\beta})$ indicates $u$'s strong preference on $i$, denoted by $r_{ui}$ for simplicity.
Each observed instance between $u$ and the interacted item $i\in \mathcal{R}_u$, 
\textit{i.e.}, $(u,i)$, can be seen as a positive label.
As for the rest unobserved interactions, \textit{i.e.}, 
$\left\lbrace (u,j)|j \notin \mathcal{R}_u \right\rbrace$, 
the probability of $(u,j)$ being negative is
\begin{equation}
P_{\text{neg}}(j|u,i) = \mathrm{sigmoid}(r_{ui} - r_{uj}),
\end{equation}
which approaches to 1 when $r_{ui}\gg r_{uj}$. In other words, when learning user preference in implicit CF, we care more about the pairwise ranking relation between an observed interaction $(u,i)\in\mathcal{R}$  and another unobserved interaction $(u,j)$, instead of absolute values of $r_{ui}$ and $r_{uj}$.
The learning objective can be formulated as minimizing following loss function~\cite{BPR}: 
\begin{equation}
L(\{\textbf{p}_u\},\{\textbf{q}_i\},\bm{\beta}) = \sum\nolimits_{(u,i)\in\mathcal{R}}
\left[
\mathbb{E}_{j\sim p_\mathrm{ns}(j|u)} 
\left[ 
-\log P_{\text{neg}}(j|u,i)
\right] 
\right] 
,
\label{Lr}
\end{equation}
where the negative instance $(u,j)$ is sampled according to a specific distribution $p_\mathrm{ns}(j|u)$.
Learning above objective is equivalent to maximizing the likelihood of observing such pairwise ranking relations $r_{ui} > r_{uj}$, 
which can be replaced by other objectives used in implicit CF problems, 
such as marginal hinge loss~\cite{ying2018graph} and binary cross-entropy loss~\cite{NCF}.

The most widely used $p_\mathrm{ns}(j|u)$
is the uniform distribution~\cite{BPR}, 
suffering from low quality of samples. 
To solve this, 
previous works~\cite{RNS,AdvIR,AdaOverSam} propose to sample much harder instances, 
containing more information.
Among them, state-of-the-arts~\cite{RNS,AdvIR} simultaneously learn a parameterized $\bar{p}_\mathrm{ns}(j|u)$ to maximize above loss function in \eqref{Lr}, based on GAN. 
Therefore, the sampled negative instance $(u,j)$ corresponds to a low $P_{\text{neg}}(j|u,i)$ and a high $r_{uj}$, 
which is generally hard for CF model to learn.
In other words, $(u,j)$ has a high probability of being positive, 
denoted as $P_{\text{pos}}(j|u,i)=1-P_{\text{neg}}(j|u,i)$.
Different choices of $p_\mathrm{ns}(j|u)$ in previous works are listed in Table~\ref{tab_complexity}. Since none of them have enough robustness to handle false negative instances, and GAN-based model structure is much more complicate, 
our goal is to propose a more robust and simplified negative sampling method.

\begin{table}[t]
	\small
	\setlength\tabcolsep{0.5pt}
	\centering
	\caption{Comparison of the proposed SRNS with closely related works, where $\text{rk}(j|u)$ is the $(u,j)$'s rank sorted by score,
		 $\text{pop}_j$ is the $j$'s item popularity, $B$ is the mini-batch size, 
		 $T$ is the time complexity of computing an instance score, 
		 $E$ is the epoch of lazy-update, and $\mathcal{F}$ denotes false negative.}
	\vspace{-5px}
	\begin{tabular}{C{65px} C{105px} C{85px}  C{70px} C{65px}}
		\toprule
		& $p_\mathrm{ns}(j|u)$ & Optimization & Time Complexity & Robustness \\ \midrule 
		Uniform~\cite{BPR} & Uniform$(\{j\notin \mathcal{R}_u\})$ & SGD~(from scratch) & $O(BT)$ & $\times$ \\ \midrule 
		NNCF~\cite{NNCF} & $\propto (\text{pop}_j)^{0.75}$ & SGD~(from scratch)  & $O(B^2T)$ &$\times$ \\ \midrule 
		AOBPR~\cite{AdaOverSam} & $\propto \exp(-\text{rk}(j|u)/\lambda)$ & SGD~(from scratch) & $O(BT)$ & $\times$ \\ \midrule 
		IRGAN~\cite{IRGAN} & learned $\bar{p}_\mathrm{ns}(j|u)$~(GAN) & REINFORCE~(pretrain) & $O(B|\mathcal{I}|T)$ & $\times$  \\ \midrule 
		AdvIR~\cite{AdvIR} & learned $\bar{p}_\mathrm{ns}(j|u)$~(GAN) & REINFORCE~(pretrain) & $O(BS_1T)$ & $\times$ \\ \midrule 
		SRNS~(proposed) & variance-based (see \eqref{argmax}) & SGD~(from scratch) & $O(\frac{B}{E}(S_1+S_2)T)$ & $\surd$\\
        \bottomrule 
   \end{tabular}
   \label{tab_complexity}
   \vspace{-10px}
\end{table}

\section{SRNS: the Proposed Method}

To improve robustness and efficiency for negative sampling in implicit CF, 
we seek for a deep understanding of different negative instances,
including false negative instances and negative instances obtained by uniform sampling or hard negative sampling. 
We then describe the proposed method based on these understandings.

\subsection{Understanding False Negative Instances}

In previous works~\cite{AdvIR,AdaOverSam}, 
the positive-label probability $P_{\text{pos}}$~(or the prediction score) is widely used as the sampling criterion, 
as it is proportional to the sample difficulty. 
Therefore, 
in Figure~\ref{fig_motivation} 
(details on setup are in Appendix~C.2), 
we have a closer look at the negative instances' distribution \textit{w.r.t.} $P_{\text{pos}}$ and further analyze the possibility of using $P_{\text{pos}}$ to discriminate true negative instances and false negative instances. 
Besides, we are also curious about the model's prediction uncertainty regarding to different negative instances, and investigate the variance of $P_{\text{pos}}$ in Figure~\ref{fig_motivation}(d).

\begin{figure}[ht]
	\centering
	\subfigure[Distribution of $P_{\text{pos}}$.]
	{\includegraphics[width=.25\textwidth]{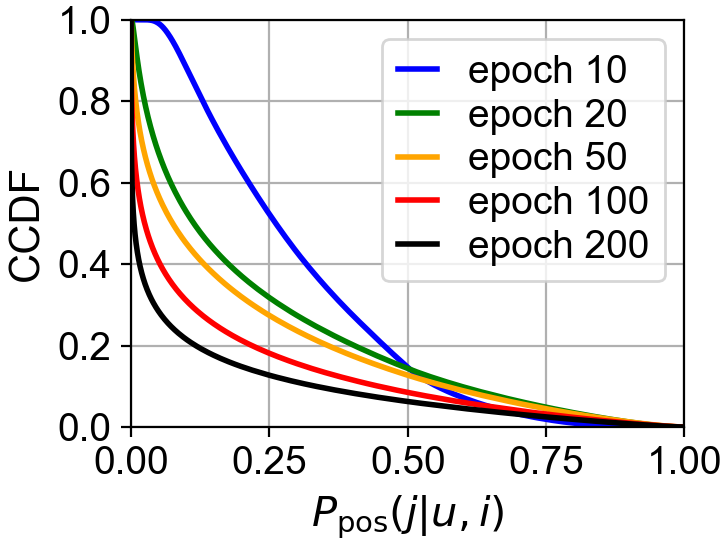}}\hfill
	\subfigure[Comparing $P_{\text{pos}}$ of neg.]{\includegraphics[width=.25\textwidth]{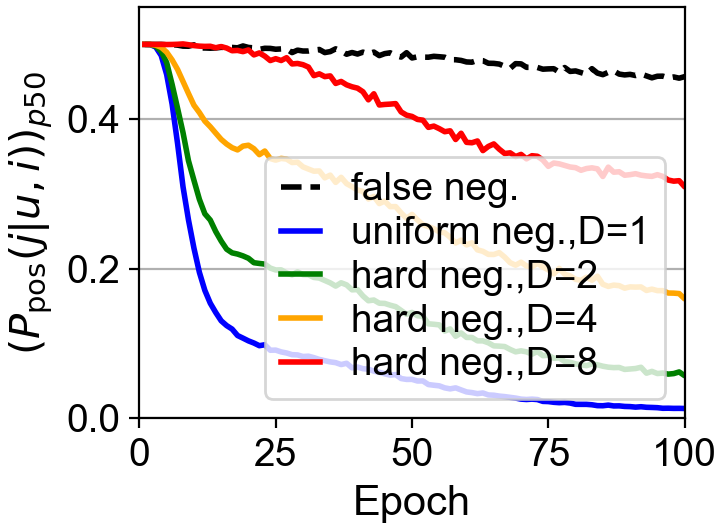}}\hfill
	\subfigure[Comparing LER.]{\includegraphics[width=.25\textwidth]{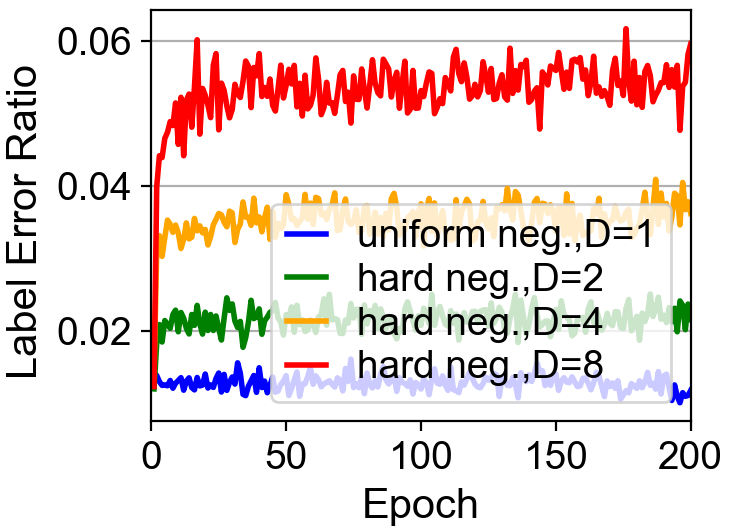}}\hfill
	\subfigure[Comparing $\frac{ \text{Std} }{ \text{Mean} }P_{\text{pos}}$.]{\includegraphics[width=.25\textwidth]{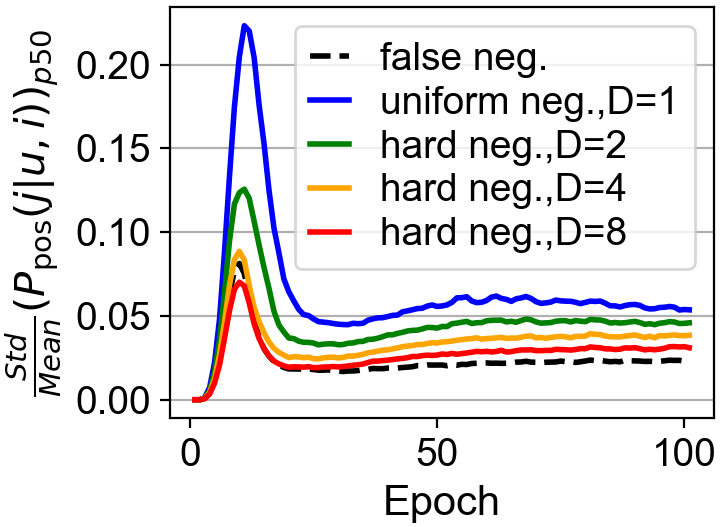}}
	\vspace{-10px}
	\caption{Analysis of negative instances on ML-100k.
		$D$: difficulty level; 
		Label Error Ratio (LER): 
		$=\nicefrac{\text{(\# of false negative samples)}}
		{ \text{ (\# of all selected negative samples) } }$;
		CCDF: complementary cumulative distribution function, $p50$: median value among a set of negative instances, 
		$\nicefrac{\text{Std}}{\text{Mean}}$: normalized variance).}
	\label{fig_motivation}
	\vspace{-12px}
\end{figure}

Based on above analysis of negative instances in implicit CF, we have following two findings:
\begin{itemize}[leftmargin=15px,noitemsep,topsep=0pt,parsep=0pt,partopsep=0pt]
\item[1)] The score distribution of negative instances is highly skew. Regardless of the training, only a few negative instances have large scores~(Figure~\ref{fig_motivation}(a)).
\item[2)] Both false negative instances and hard negative instances have large scores~(Figure~\ref{fig_motivation}(b)), making it hard to discriminate them~(harder negative samples are more likely be false negative, Figure~\ref{fig_motivation}(c)). However, the false negative instances have lower prediction variance comparatively~(Figure~\ref{fig_motivation}(d)).
\end{itemize}
The first finding demonstrates the potential of just capturing a part of the full distribution corresponding to those large-scored negative instances, which are more likely to be high-quality.
Similar observations have also been discussed in graph representation learning, suggesting a skewed negative sampling distribution that focuses on hard ones~\cite{yang2020understanding,NSCaching}.

As for the second finding, it provides us a reliable way of measuring sample quality based on prediction variance, sharing the same intuition with \cite{ActiveBias} that improves stochastic optimization by emphasizing high variance samples.
Specifically, we prefer those negative samples with both large scores and high variances, avoiding false negative instances that always have large scores over many iterations of training. 
In terms of robustifying negative sampling process, none of above works in implicit CF and other domains have tackled this problem, except for a simple workaround that only selects hard negative samples but avoids the hardest ones~\cite{ying2018graph,NSCaching}.

\subsection{SRNS Method Design}

As above, 
on the one hand, we are motivated to use a small amount of memory for each user, storing hard negative instances that have large potential of being high-quality. This largely simplifies the model structure, by focusing on a partial set of instances, which thus improves efficiency.
On the other hand, we propose a variance-based sampling strategy to effectively obtain samples that are both reliable and informative.
Our simplified and robust negative sampling~(SRNS) approach addresses the remaining challenges on model efficiency and robustness.
Algorithm~\ref{algo} shows the implicit CF learning framework, \textit{i.e.}, minimizing loss function in \eqref{Lr}, based on SRNS.

\vspace{-5px}
\begin{algorithm}[ht]
\caption{The proposed Simplified and Robust Negative Sampling~(SRNS) method.}
\small
\label{algo}
\SetKwInOut{Input}{Input}\SetKwInOut{Output}{Output}
\Input {Training set $\mathcal{R}=\{(u,i)\}$, embedding dimension $F$, scoring function $r$ with learnable parameter $\bm{\beta}$, and memory $\{\mathcal{M}_u| u\in\mathcal{U}\}$, each with size $S_1$;}
\Output {Final user embeddings $\{\mathbf{p}_u|u\in\mathcal{U}\}$ and item embeddings $\{\mathbf{q}_i| i\in\mathcal{I}\}$, and $r$;}
Initialize $\{\mathbf{p}_u| u\in\mathcal{U}\}$ and $\{\mathbf{q}_i| i\in\mathcal{I}\}$, $\bm{\beta}$ and $\{\mathcal{M}_u| u\in\mathcal{U}\}$; \\
\For{$t = 1,2,...,T$}{
Sample a mini-batch $\mathcal{R}_{batch} \in \mathcal{R}$ of size $B$; \\
\For{each $(u,i)\in \mathcal{R}_{batch}$}{
Get the candidate items from $u$-related memory $\mathcal{M}_u$; \\
Sample the item $j$ from $\mathcal{M}_u$, based on the variance-based sampling strategy~\eqref{argmax}; \label{line_s}\\
Uniformly sample $S_2$ items from $\{k|k\notin \mathcal{R}_u\}$~($\bar{\mathcal{M}}_u$), and merge with original $\mathcal{M}_u$; \label{line_u1}\\
Update $\mathcal{M}_u$ based on the score-based updating strategy~\eqref{eq:score};\label{line_u2}\\
Update embeddings $\{\mathbf{p}_u,\mathbf{q}_i,\mathbf{q}_j\}$ and parameters $\bm{\beta}$ based on gradient \textit{w.r.t.} $L$~\eqref{Lr}.
}
}
\end{algorithm} 
\vspace{-5px}

The learning process of the SRNS is carried out in mini-batch mode, and alternates between two main steps. 
First, according to the high-variance based criterion, a negative instance for each training instance $(u,i)$ is sampled from $u$'s memory $\mathcal{M}_u$~(line~\ref{line_s}), which already stores $S_1$ candidates with high potential. To improve efficiency, all positive instances of a same user $u$ is designed to share one memory $\mathcal{M}_u$.
Second, as the model is constantly changing during the training process, $\mathcal{M}_u$ requires a dynamic update so as to keep track of the promising candidates for negative sampling. Specifically, this is completed by first extending it into $\mathcal{M}_u \cup \bar{\mathcal{M}}_u$ with additional $S_2$ instances that was uniformly sampled~(line~\ref{line_u1}), and then choosing $S_1$ hard candidates to obtain a new $\mathcal{M}_u$~(line~\ref{line_u2}). 
A similar two-step scheme is adopted by a related work that focuses on negative sampling for knowledge graph embeddings~\cite{NSCaching}. 
However, unlike SRNS leveraging the instance's variance in the sampling step, it uniformly chooses an instance from memory, which cannot enhance model's robustness effectively.

\textbf{Score-based memory update.}
In this part, we propose a memory-based model to simply capture the dynamic distributions of true negative instances.
Specifically, a memory $\mathcal{M}_u=\{(u,k_1),...,(u,k_{S_1})\}$ of size $S_1$ is assigned to each user $u$, 
storing the negative instances that are available to $u$ in sampling. 
To ensure only those informative instances are maintained, 
we design a score-based strategy to dynamically update the $\mathcal{M}_u$, which tends to involve more hard negative instances. 
For an extended memory that merges the old $\mathcal{M}_u$ and a set of uniformly sampled instances $\bar{\mathcal{M}}_u$, \textit{i.e.}, 
$\mathcal{M}_u \cup \bar{\mathcal{M}}_u$, 
the new $\mathcal{M}_u$ is updated by sampling $S_1$ instances according to the following probability distribution:
\begin{equation}
\bar{\Psi}(k|u,\mathcal{M}_u \cup \bar{\mathcal{M}}_u) 
= \exp (\nicefrac{r_{uk}}{\tau}) / \sum\nolimits_{k' \in \mathcal{M}_u \cup \bar{\mathcal{M}}_u} \exp (\nicefrac{r_{uk'}}{\tau}),
\label{eq:score}
\end{equation}
where a lower temperature $\tau\in(0,+\infty)$ would make $\bar{\Psi}$ focus more on large-scored instances.

\textbf{Variance-based sampling.}
As we have demonstrated in finding 2, oversampling hard negative instances may increase the risk of introducing false negatives, 
making above score-based updating strategy less robust.
Motivated by the observed low-variance characteristic of false negatives, we propose a robust sampling strategy that can effectively avoid this noise by favouring those high-variance candidates.
Given a positive instance $(u,i)$ and $u$'s memory $\mathcal{M}_u$, 
for each candidate $(u,k) \in \mathcal{M}_u$, 
we maintain $P_{\text{pos}}(k|u,i)$ values at $t$th training epoch as $[P_{\text{pos}}(k|u,i)]_t$.
The proposed variance-based sampling strategy chooses the negative instance $(u,j)$ from $\mathcal{M}_u$ by:
\begin{equation}
j = \mathop{\arg\max}
\nolimits_{k \in \mathcal{M}_u} P_{\text{pos}}(k|u,i) 
+ \alpha_t \cdot \mathrm{std}[P_{\text{pos}}(k|u,i)].
\label{argmax}
\end{equation}
Note that we also consider the instance difficulty,
i.e., $P_{\text{pos}}(k|u,i)$, 
to ensure the informativeness of sampled negative instances, with a hyper-parameter $\alpha_t$ controlling the importance of high-variance at the $t$-th training epoch. When $\alpha_t=0$, our proposed sampling approach degenerates into a difficulty-only strategy that follows the similar idea as previous works~\cite{RNS,AdvIR,AdaOverSam}.
Since all instances tend to have high variance at an early training stage, the variance term should not be weighted too much.
Therefore, we expect a ``warm-start'' setting of $\alpha_t $ that reduces the influence of prediction variance at first and then gradually strengthens it~(details of $\alpha_t $ will be discussed in Section~\ref{sec:exp1}).

\textbf{Bootstrapping.}
In Algorithm~\ref{algo}, 
false negative instances are identified by checking their prediction variance. 
However, 
this assumes that the CF model has some discriminative ability. 
There is an important observation that deep models can memorize easy training instances first and gradually adapt to hard instances~\cite{memorization,coteaching,zhang2016understanding}.
Fortunately, 
we also observe such memorization effect for deep CF models (see Section~\ref{sec:exp1}), 
which means that the false negative instances among unlabeled data are generally more difficult and may not be memorized at an early stage.
In other words, 
SRNS can be self-boosted by first learning to discriminate those easy negative instances and then tackling the rest real hard ones with the help of variance-based criterion.

\subsection{Complexity Analysis}

Here, we analyze the time complexity of SRNS (Algorithm~\ref{algo})
and compare it with related negative sampling approaches in Table~\ref{tab_complexity}.
Compared with a uniform sampling approach~\cite{BPR}, the main additional cost comes from score-based memory update and variance-based sampling. 
The former requires to compute scores of $S_1+S_2$ candidates for each positive instance and sample $S_1$ of them according to a normalized distribution $\bar{\Psi}$ that is based on computed scores. Thus the time complexity is $O((S_1+S_2)T)$, where $T$ denotes the operation count of score computation. As for the latter, computing $\mathrm{std}[P_{\text{pos}}(k|u,i)]$ of each candidate and choosing the final negative instance in \eqref{argmax} take $O(S_1)$. 
Thus, 
the cost is $O((S_1+S_2)T)$ for each positive instance, 
which can be reduced to $O((S_1+S_2)T/E)$ using lazy-update every $E$ epochs.
Model parameters in CF consists of two parts, 
\textit{i.e.}, embeddings and scoring function parameters, and the former is generally much larger.
Specifically, SRNS's model complexity is about $(|\mathcal{U}|+|\mathcal{I}|)F$, which can double in those GAN-based state-of-the-arts~\cite{AdvIR,IRGAN}.
As in Table~\ref{tab_complexity}, SRNS is not only more simplified~(in terms of time complexity),
but also can be easily trained from scratch.

\section{Experiments}
We first conduct controlled experiments with synthetic noise, 
so as to investigate SRNS's robustness to false negative instances~(Section~\ref{sec:exp1}).
Then, we evaluate the SRNS's performance on the implicit CF task, based on real data experiments~(Section~\ref{sec:exp2}).

\subsection{Experimental Settings}\label{exp_setting}
\textbf{Dataset.}
Table~\ref{tab:data} summarizes datasets used for experiments,
which are popularly used in the literature~\cite{geng2015learning,NCF,AdvIR,IRGAN}.
We use ML-100k and Ecom-toy for synthetic data experiments and do a 4:1 random splitting for train/test data.
To simulate the noise, 
we randomly select 50\%/25\%  of groundtruth records in the test set of ML-100k/Ecom-toy. The selected records can be regarded as false negative instances during training, denoted as $\mathcal{F}$.
As for real data experiments, we use the rest three datasets and adopt leave-one-out strategy, \textit{i.e.}, holding out users' most recent records and second to the last records~(sorted \textit{w.r.t.} time-stamp) as the test set and validation set, respectively~\cite{NCF,BPR}.

\begin{table}[ht]
	\small
	\setlength\tabcolsep{0.5pt}
	\centering
	\caption{Statistics of datasets.}
	\begin{tabular}{C{60px} C{60px} C{40px} C{40px}  C{40px} C{40px} C{40px} C{60px} }
	\toprule
	Category& Dataset & User & Item & Train & Validation & Test & $\mathcal{F}$~(Noise) \\ 
	\midrule
	\multirowcell{2}{Synthetic \\ noisy dataset} & Movielens-100k &942&1,447&44,140&-&11,235&5,509  \\
    &Ecommerce-toy &1,000&2,000&60,482&-&14,612&3,246 \\	
	\midrule
	\multirowcell{3}{Real-world \\ dataset} & Movielens-1m &6,028&3,533&563,186&6,028&6,028&- \\
	& Pinterest &55,187&9,916&1,390,435&55,187&55,187&- \\
	&Ecommerce &66,450&59,290&1,625,006&66,441&66,450&- \\ \bottomrule
	\end{tabular}
	\label{tab:data}
\end{table}

\textbf{Baselines.}
We compare SRNS with three types of methods listed in Table~\ref{tab_complexity}. 
First, for methods using a fixed negative sampling distribution, we choose Uniform~\cite{BPR} and NNCF~\cite{NNCF}.
Second, for methods based on hard negative sampling, we choose AOBPR~\cite{AdaOverSam}, IRGAN~\cite{IRGAN}, RNS-AS~\cite{RNS} and AdvIR~\cite{AdvIR}, where the last three are GAN-based state-of-the-arts.
Finally, we also compare with a non-sampling approach ENMF~\cite{ENMF} that regards all the unlabeled data as negative labels.

\textbf{Hyper-parameter and optimizer.}
For better performance, we mainly use GMF~\cite{NCF} as the scoring function $r$, but also experiment on another popular choice, \textit{i.e.}, a MLP with sigmoid activation~(Section~\ref{sec:exp2}).
Hyper-parameters of SRNS and baselines are carefully tuned according to validation performance~(details are in Appendix~B.4).
For all experiments, Adam optimizer is used and the mini-batch size is 1024.
Specifically, we run each synthetic data experiment 400 epochs and repeat five times.
As for real data experiments, we conduct the standard procedure~\cite{NNCF,wang2019neural}, running 400 epochs and terminating training if validation performance does not improve for 100 epochs.

\textbf{Experimental setup.}
In the implicit CF, the model is evaluated by testing if it can generate a better ranked item list $\mathcal{S}_u$ for each user $u$.
In the synthetic case, $\mathcal{S}_u$ contains $u$'s test items $\mathcal{G}_u$ and the rest items that are not interacted by $u$. While in the real-world case with much larger item count $|\mathcal{I}|$, we follow a common strategy~\cite{NCF,koren2008factorization} to fix the list length $|\mathcal{S}_u|$ as 100, by randomly sampling $100-|\mathcal{G}_u|$ non-interacted items. 
We measure $\mathcal{S}_u$'s performance by \textit{Recall} and \textit{Normalized Discounted Cumulative Gain}~(NDCG). Specifically, 
$Recall@k(u)=|\mathcal{S}_u(k) \cap \mathcal{G}_u|/|\mathcal{G}_u|$, $NDCG@k(u)=\sum_{i \in \mathcal{S}_u(k) \cap \mathcal{G}_u} 1 / \log_2(p_i+1)$, where $\mathcal{S}_u(k)$ denotes truncated $\mathcal{S}_u$ at $k$ and $p_i$ denotes $i$'s rank in  $\mathcal{S}_u$.
Comparatively, NDCG accounts more for the position of the hit by assigning higher scores to hits at top ranks and $NDCG@1(u)$=$Recall@1(u)$. We choose a rather small $k$ in $\{1,3\}$, which matters more in applications.
The final report Recall/NCDG is the average score among all test users.

\subsection{Synthetic Noise Experiments} 
\label{sec:exp1}
Synthetic false negative instances are simulated by flipping labels of test records~($\mathcal{F}$ in Table~\ref{tab:data}).
To manually inject this noise, we constantly feed a false negative into each user's memory $\mathcal{M}$ during sampling process.
We control this impact by varying the size of available false negative instances in different experiments, randomly sampling $\sigma\times100$~(\%) from $\mathcal{F}$~($\sigma\in[0,1]$).
Note that $\sigma=0$ indicates an ``ideal'' case where $\mathcal{M}$ is not influenced by $\mathcal{F}$.
In these experiments, we fix the memory size $S_1$ as 20~(details on setup are in Appendix~C.3).

\begin{figure}[t]
\centering
\subfigure[N@3, ML-100k]{\includegraphics[width=.245\textwidth]{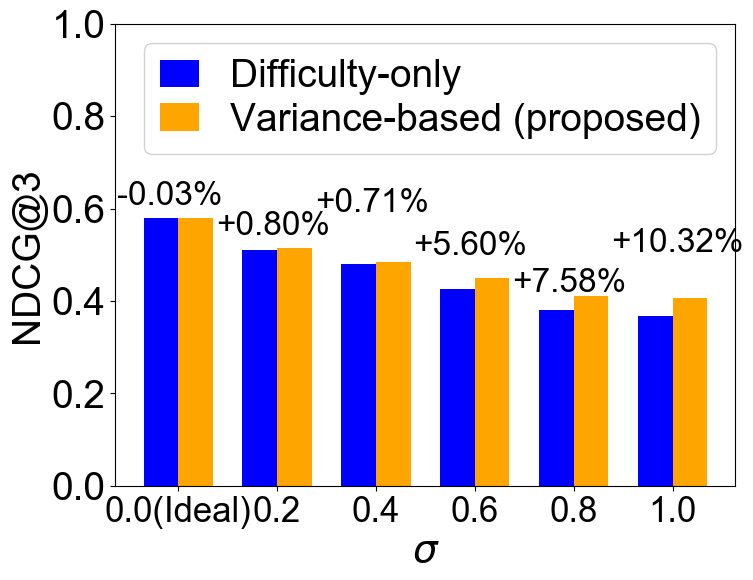}}\hfill
\subfigure[R@3, ML-100k]{\includegraphics[width=.245\textwidth]{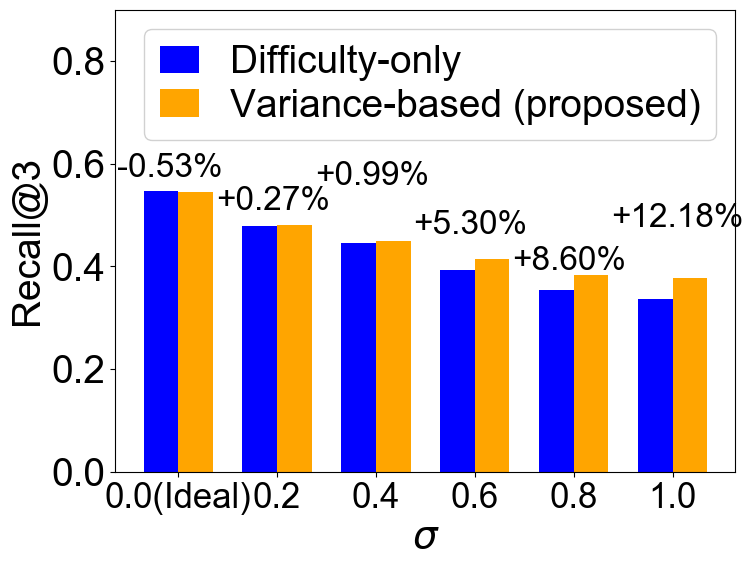}}\hfill
\subfigure[N@3, Ecom-toy]{\includegraphics[width=.25\textwidth]{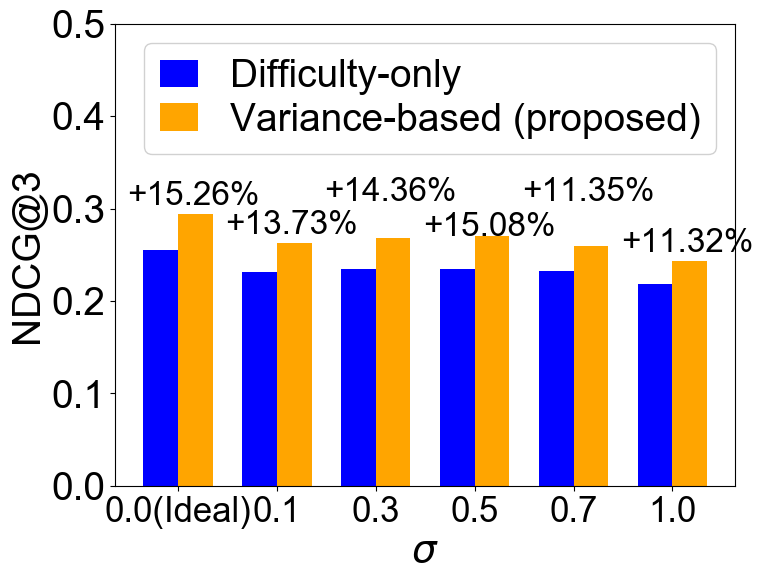}}\hfill
\subfigure[R@3, Ecom-toy]{\includegraphics[width=.25\textwidth]{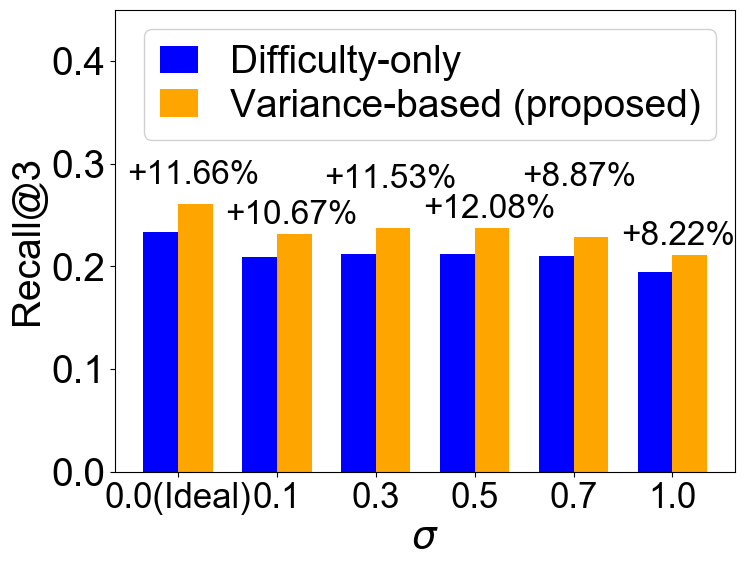}}\hfill
\vspace{-8px}
\caption{Average test Recall/NDCG of SRNS with different $\sigma$ on two synthetic datasets over the last 50 epochs. Two sampling strategies are compared, \textit{i.e.}, difficulty-only vs. variance-based~(proposed).  }	
\label{fig_synthetic1}
\vspace{-10px}
\end{figure}

\textbf{Sampling criterion.}
We first investigate if the high-variance based criterion in SRNS can indeed identify true negative instances which are of high-quality, by comparing with a difficulty-only strategy~(\textit{i.e.}, weight $\alpha_t = 0$).
Figure~\ref{fig_synthetic1} shows comparison results \textit{w.r.t.} test Recall and NDCG, 
under different levels of noisy supervision~($\sigma$).
Although increasing noisy level can harm model's performance, 
we can observe a consistent improvement of variance-based strategy with different $\sigma$.

\begin{figure}[ht]
\centering
\vspace{-5px}
\subfigure[$\sigma=1$, ML-100k]{\includegraphics[width=.25\textwidth]{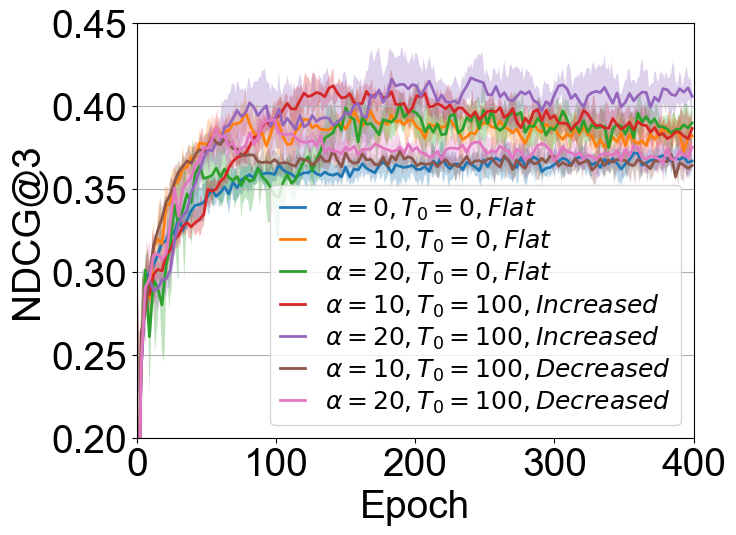}}\hfill%
\subfigure[$\sigma=0.5$, Ecom-toy]{\includegraphics[width=.25\textwidth]{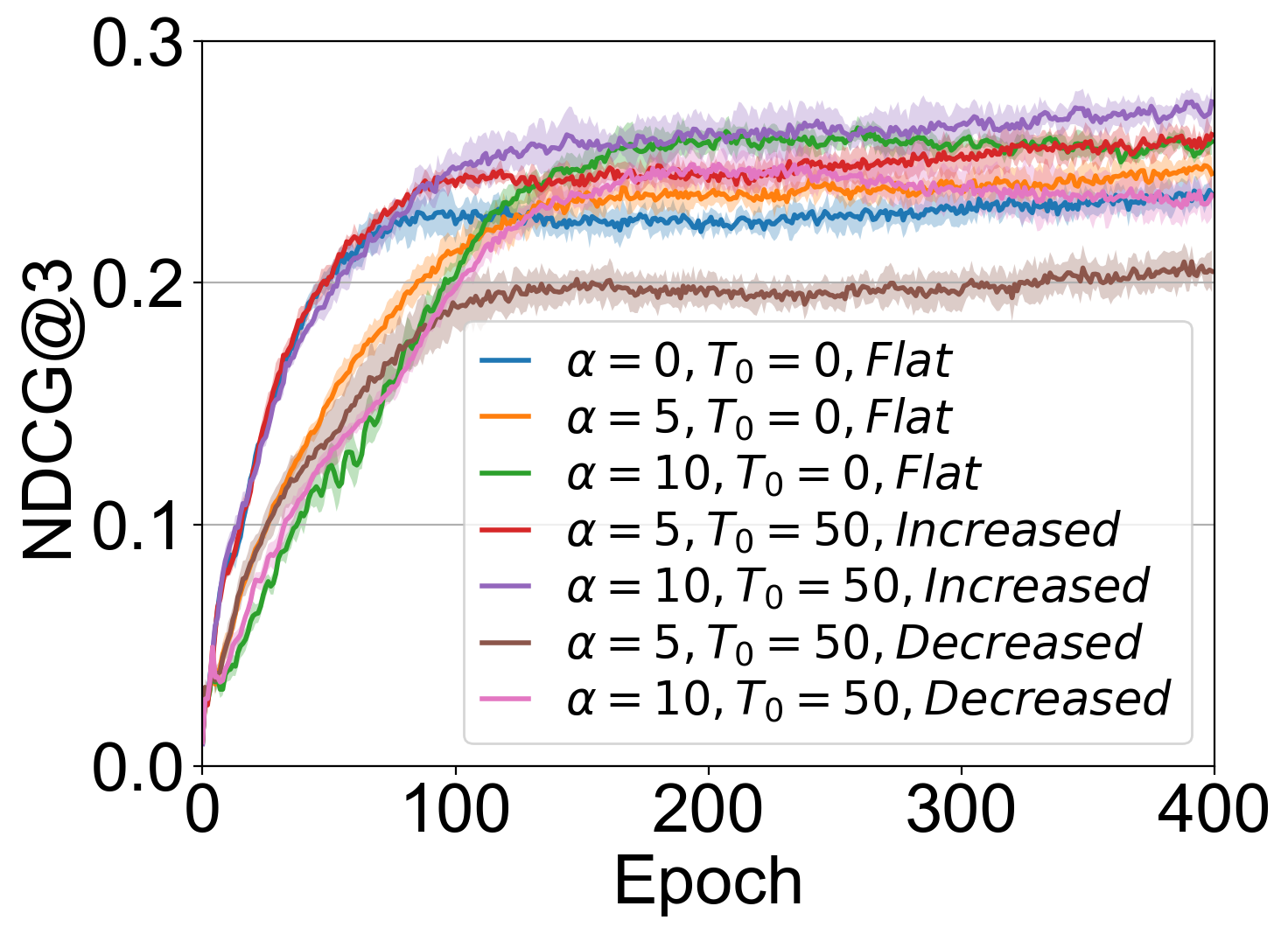}}\hfill%
\subfigure[$\sigma=1$, ML-100k]{\includegraphics[width=.25\textwidth]{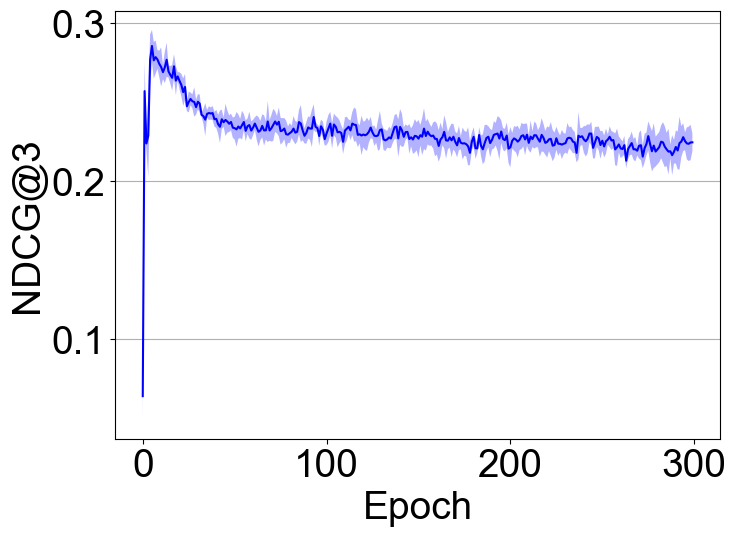}}\hfill
\subfigure[$\sigma=1$, Ecom-toy]{\includegraphics[width=.25\textwidth]{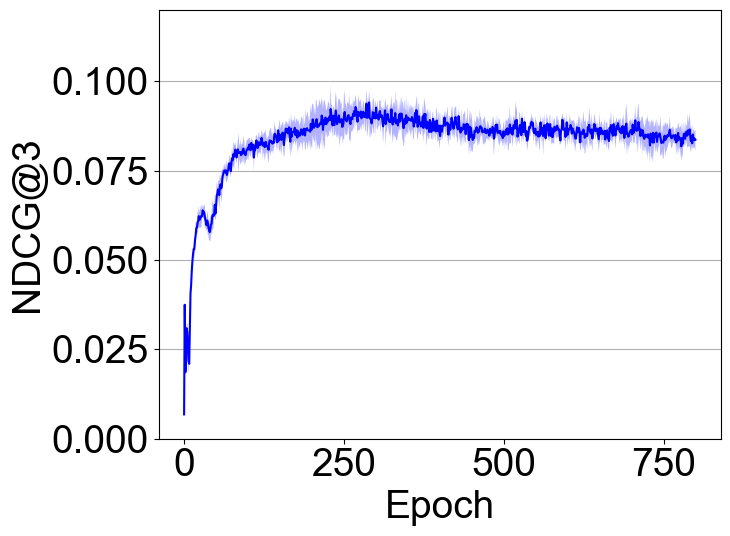}}
\vspace{-8px}
\caption{(a)-(b) Test NDCG vs. number of epochs on two datasets, with the error bar for STD highlighted as a shade. (c)-(d) Memorization effect of the CF model under extremely noisy supervision. }	
\label{fig_synthetic2}
\end{figure}

\textbf{Warm-start.}
Motivated by \cite{coteaching}, we propose to linearly increase the value of $\alpha_t$ as epoch number $t$ increases. Specifically, $\alpha_t = \alpha \cdot \min(t/T_0,1)$, where $T_0$ denotes the threshold of stopping increase.
In Figure~\ref{fig_synthetic2}(a)-(b), we compare this increased setting of $\alpha_t$ with another two competitor, 
\textit{i.e.}, $\alpha_t = \alpha$~(flat) and $\alpha_t = \alpha \cdot \max(1-t/T_0,0)$~(decreased).
It can be clearly observed that the increased setting of $\alpha_t$ performs better than others, 
as the former can better leverage variance-based criterion after false negative instances become more stable.  

\textbf{Bootstrapping.}
Finally we demonstrate SRNS's self-boosting capability, by illustrating the memorization effects of CF models in Figure~\ref{fig_synthetic2}(c)-(d). To ensure a clear observation, we inject much intenser noise during sampling process, by extending $\mathcal{F}$ to 100\% and 40\%  of the original test set on ML-100k and Ecom-toy, respectively.
Under extremely noisy supervision~($\sigma=1$), though sampling based on difficulty only~($\alpha_t = 0$), the model's test NDCG first reaches a high level and then gradually decreases, indicating that it can avoid the impact of false negative instances at an early stage.

\subsection{Real Data Experiments} \label{sec:exp2}

\begin{table}[b]
\vspace{-10px}
	\small
	\renewcommand{\arraystretch}{0.8}
	\setlength\tabcolsep{0.5pt}
	\centering
	\caption{Performance comparison \textit{w.r.t.} test NDCG and Recall on three datasets.
		The last row shows relative improvement in percentage compared with the second best.}
	\begin{tabular}{C{60px} C{50px} C{30px}  C{30px} C{30px} C{30px} C{30px} C{30px}  C{30px} C{30px} C{30px} }
		\toprule
		\multirowcell{2}{Category} & \multirow{2}{*}{Method} & \multicolumn{3}{c}{Movielens-1m} & \multicolumn{3}{c}{Pinterest} & \multicolumn{3}{c}{Ecommerce} \\ 
		& & N@1  & N@3  & R@3  & N@1  & N@3  & R@3 & N@1  & N@3  & R@3\\ \midrule
		Non-sampling &ENMF        &\underline{0.1846} &\underline{0.3021} &\underline{0.3882} &0.2594 &0.4144 &0.5284 &0.1317 &0.2095 &0.2670   \\ \midrule
		\multirowcell{2}{Fixed Dist. \\ Sampling} & Uniform      &0.1744 &0.2846 &0.3663 &0.2586 &0.4136 &0.5276 &0.1265 &0.2057 &0.2640 \\ 
		& NNCF &0.0829&0.1478&0.1971&0.2292&0.3699&0.4735&0.0833&0.1420&0.1855 \\ \midrule
		\multirowcell{4}{Hard \\Negative \\Sampling} &AOBPR       &0.1802 &0.2905 &0.3728 &0.2596 &0.4165 & 0.5319 &0.1293 &0.2108 &0.2710 \\ 
		& IRGAN        &0.1755 &0.2877 &0.3708 &0.2587 &0.4143 &0.5282 &0.1275 &0.2065 &0.2648 \\ 
		&RNS-AS       &{0.1823} &0.2932 &0.3754 &\underline{0.2690} &0.4233 &0.5359  &0.1335 &0.2131 &0.2714 \\ 
		&AdvIR       &0.1790 &{0.2941} &{0.3792} &0.2689 &\underline{0.4235} &\underline{0.5363} &\underline{0.1357} &\underline{0.2141} &\underline{0.2719}  \\ \midrule
		\multirowcell{2}{Proposed} &SRNS        &\textbf{0.1933} &\textbf{0.3070} &\textbf{0.3912} &\textbf{0.2891} &\textbf{0.4391} &\textbf{0.5486} &\textbf{0.1471} &\textbf{0.2256} &\textbf{0.2833}  \\ 
		&    &4.71\% &1.62\%  &0.77\% &7.47\% &3.68\% &2.29\% &8.40\% &5.37\% &4.19\%  \\ \bottomrule
	\end{tabular}
	\label{tab_comp}
\end{table}

\textbf{Performance comparison.}
As shown in Table~\ref{tab_comp}, we compare SRNS with seven baselines \textit{w.r.t.} test NDCG and Recall on three real-world datasets.
As can be seen, SRNS consistently outperforms them, achieving a relative improvement of 4.71$\sim$8.40\% \textit{w.r.t.} NDCG@1. This indicates that SRNS can sample high-quality negative instances and thus helps to learn a better CF model that ranks items more accurately.
Specifically, we have following three observations. 
First, among all baselines, hard negative sampling approaches perform more competitively. By considering both informativeness and reliability of negative instances, our SRNS outperforms two state-of-the-art baselines, \textit{i.e.}, RNS-AS and AdvIR, that generate hard negatives based on adversarial sampling.
Second, approaches using a fixed sampling distribution perform poorly, especially NNCF that directly adopts a power distribution based on item popularity.
Third, by improving sample quality, sampling-based approaches can be more effective than the non-sampling counterpart that models the whole unlabeled data. For example, ENMF performs worse than RNS-AS and AdvIR on Pinterest and Ecom.

\begin{figure*}[t]
\centering
\vspace{-5px}
\subfigure[Time, ML-1m]
{\includegraphics[width=.255\textwidth]{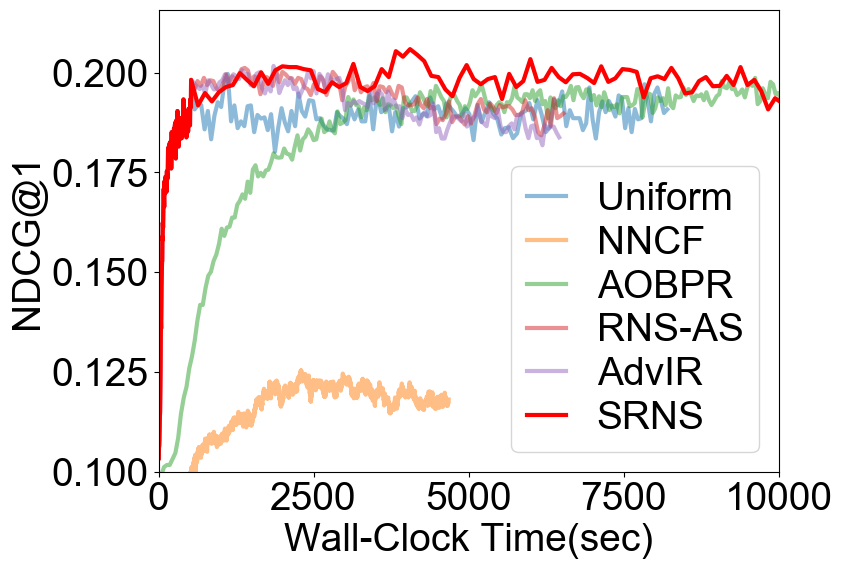}}
\subfigure[Time, Pinterest]
{\includegraphics[width=.255\textwidth]{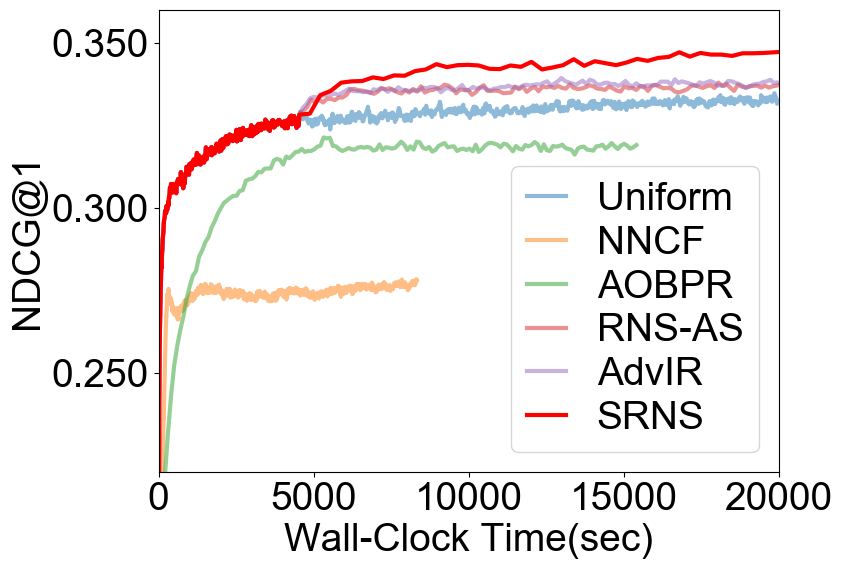}}
\subfigure[Time, Ecom]
{\includegraphics[width=.255\textwidth]{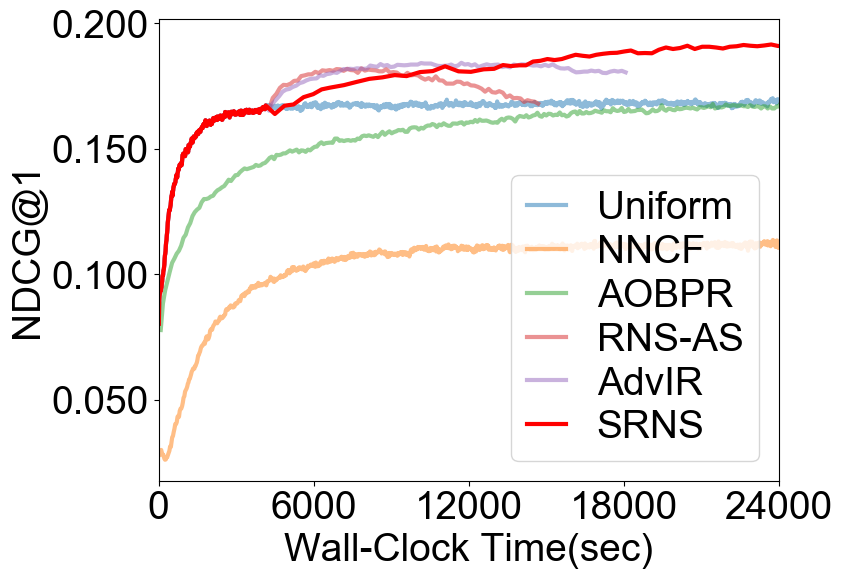}}
\subfigure[$S_1$, ML-1m]
{\includegraphics[width=.235\textwidth]{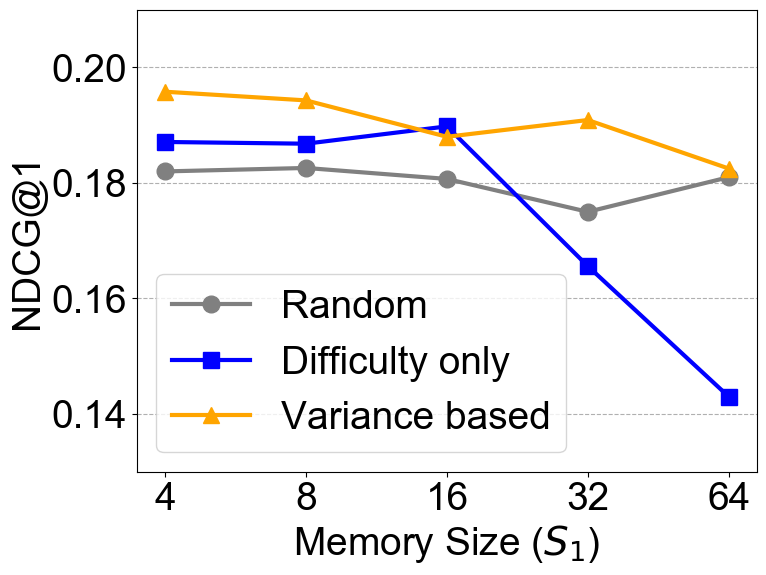}}
\vfill\vspace{-5px}
\subfigure[$S_1$, Pinterest]
{\includegraphics[width=.235\textwidth]{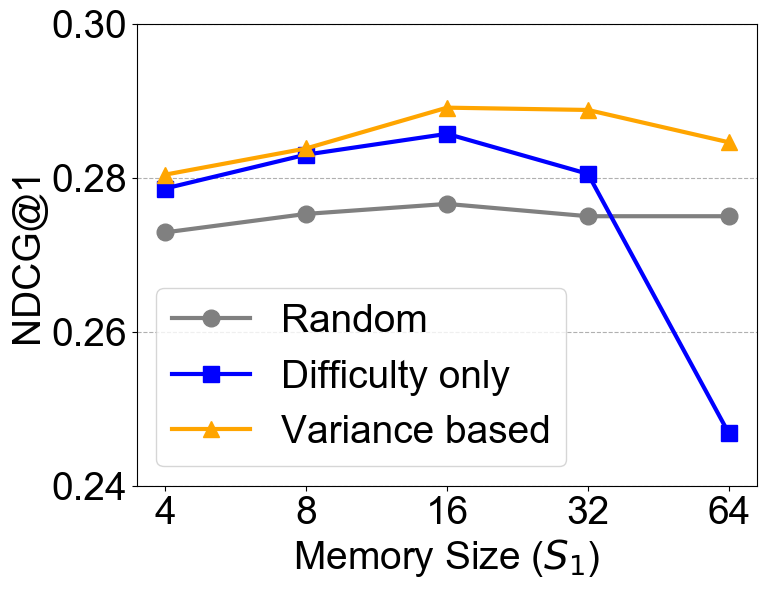}}\hspace{+0.5em}
\subfigure[$S_1$, Ecom]
{\includegraphics[width=.235\textwidth]{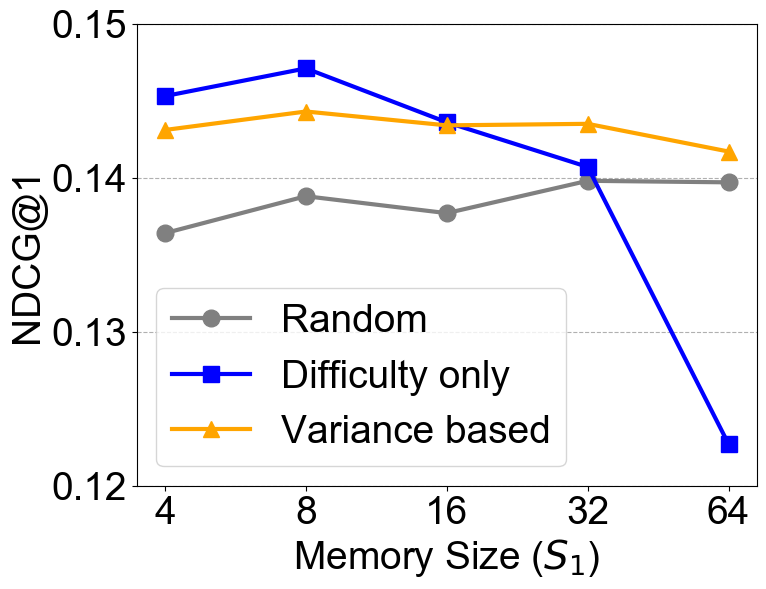}}\hspace{+0.5em}
\subfigure[$r$, ML-1m]
{\includegraphics[width=.235\textwidth]{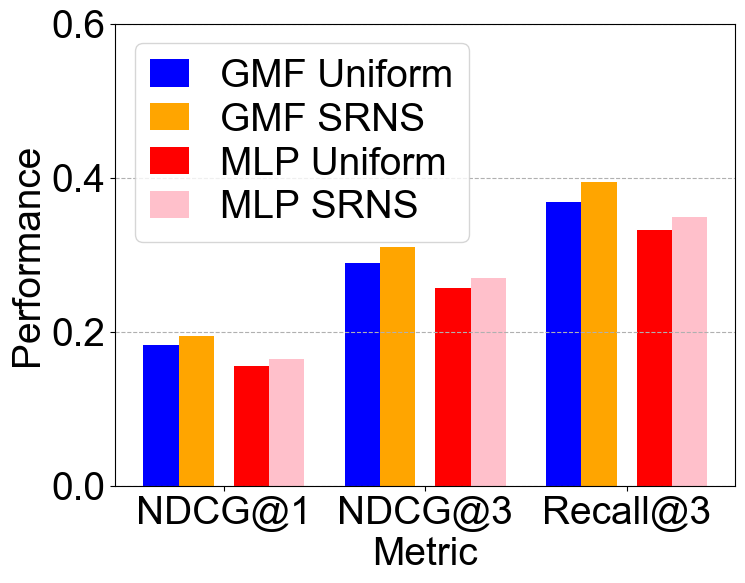}}\hspace{+0.5em}
\subfigure[$r$, Ecom]
{\includegraphics[width=.235\textwidth]{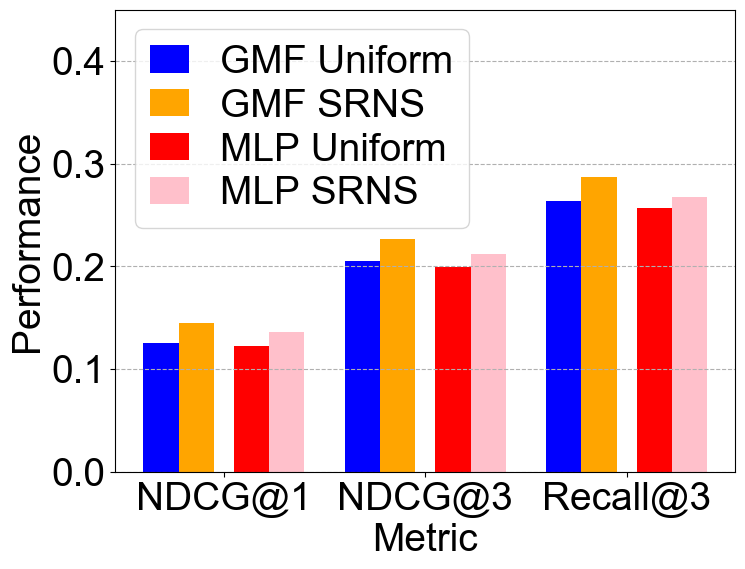}}
\vspace{-8px}
\caption{(a)-(c) Validation NDCG vs. wall-clock time~(in seconds) on three datasets. (d)-(f) Test NDCG vs. SRNS's memory size $S_1$, using different sampling strategies on three datasets. (g)-(h) Test NDCG and Recall of Uniform and SRNS, using two $r$~(GMF and MLP), on ML-1m and Ecom.}	
\label{fig_time}
\vspace{-10px}
\end{figure*}

Besides effectiveness, we also compare performance in terms of efficiency, 
by illustrating validation NDCG vs. wall-clock time in Figure~\ref{fig_time}(a)-(c). 
We observe that SRNS can converge much faster and is more stable than RNS-AS and AdvIR that use GAN based structure. 
For fair efficiency comparison, 
here we also start training SRNS from the same pretrained model as in RNS-AS and AdvIR. 

\textbf{Robustness of variance-based sampling.}
With the score-based updating strategy, increasing memory size $S_1$ makes SRNS more prone to the false negatives.
Therefore, we further test robustness of the variance-based sampling~(in \eqref{argmax}), by evaluating performance under different $S_1$. 
As illustrated in Figure~\ref{fig_time}(d)-(f), variance-based SRNS~(orange line) performs stably, indicating that emphasizing high-variance can reliably obtain high-quality samples. 
Comparatively, difficulty-only strategy~($\alpha_t=0$, blue line) suffers from dramatic degradation~($S_1$=32 or 64).
Another strategy for avoiding false negatives is to randomly select a sample from memory~\cite{NSCaching}, which performs less effectively than our approach.
Note that the necessity of variance-based sampling depends on the specific real-world data and, for example, Ecom may not need this \textit{w.r.t.} overall best NDCG@1~(Figure~\ref{fig_time}(f)).
Generally, our SRNS is flexible enough to switch between different situations, by controlling importance of variance-based sampling criterion~(\textit{i.e.}, $\alpha_t$).

\textbf{Varying scoring function.}
Finally we test SRNS's effectiveness on different $r$ including GMF and MLP~\cite{NCF}.
As illustrated in Figure~\ref{fig_time}(g)-(h), we observe similar performance improvement of SRNS over Uniform~\cite{BPR} when using the above two $r$, indicating SRNS's capability of combining with different $r$. We are also interested in exploring more choices like GNN-based $r$~\cite{ying2018graph} in future study.
Note that embedding dimension $F$ is set as 32~(ML-1m), 16~(Pinterest) and 8~(Ecom), respectively, as we observe similar results with different $F\in\{8,16,32,64\}$~(details are in Appendix~C.4).

\section{Conclusion}
In this paper, we propose a simplified and robust negative sampling approach SRNS for implicit CF, which can efficiently sample true negative instances that are of high-quality.
Motivated by the empirical evidence on different negative instances, our score-based memory design and variance-based sampling criterion achieve efficiency and robustness, respectively, in negative sampling.
Experimental results on both synthetic and real-world datasets demonstrate SRNS's robustness and superiorities.
Finally,
one interesting future works would be
studying the theoretical convergence guarantee of the proposed method.
We will attempt to address this issue by learning from  
importance sampling methods \cite{ActiveBias,zhao2015stochastic} in stochastic optimization.

\clearpage

\bibliographystyle{plain}
\bibliography{ref}

\clearpage

\setcounter{table}{3}
\setcounter{figure}{4}

\appendix

\section{Comparison Between Different Approaches}\label{appendix:difference}

\subsection{General Machine Learning Approaches}
Learning an implicit CF model from the positive-only data is also related to Positive-Unlabeled~(PU) learning and learning from noisy labels, as the rest unobserved instances are unlabeled and noisy.
Motivated by these general machine learning approaches,
this paper formulates the negative sampling problem as efficient learning from unlabeled data with the presence of noisy labels, and pays more attention on those true negative instances hidden inside the massive unlabeled data.
The following table and review on literatures discuss the differences between different approaches that can be adapted for this problem.

\begin{table}[ht]
	\centering
	\small
	\begin{tabular}{C{70px} C{100px} C{100px} C{80px}}
		\toprule
		              Approaches               & Learning from Positive-Unlabeled Data           & Learning from Noisy Labels                                        & Negative Sampling                                      \\ \midrule
		  Positive/Negative  Class Prior   & known or estimated from data~\cite{penL14PE,KM4MPE} & unknown                                                                                              & unknown                                                \\ \midrule
		     Assumption on Unlabeled Data      & positive or 
			 negative labels \cite{weightedPU,nnPU}                     & negative labels 
			  with noise \cite{mentornet,L2Reweight}           & unobserved \cite{word2vec,deepwalk,BPR} 
		                        \\ \midrule
		Handling Uncertainty of Unlabeled Data & minimizing the empirical risk estimator~\cite{unbiasPU,nnPU}  & manually designing~\cite{curriculum,selfpaced} or automated learning the instance weight~\cite{mentornet,L2Reweight,metaweight}  & sampling unobserved instances as negative labels~\cite{word2vec,deepwalk,BPR}  \\ \bottomrule
	\end{tabular}
\end{table}

\textbf{Learning from Positive-Unlabeled Data.}
Since implicit feedback data contain positive instances only, the implicit CF problem is also related to learning from positive-unlabeled~(PU) data.
PU learning formulates the problem as a binary classification, accounting for the fact that both positive and negative labels exist in the unlabeled data~\cite{unbiasPU,weightedPU,nnPU}. 
However, it normally requires an accurate estimation of the class-prior, which is challenging in real-world data~\cite{penL14PE,KM4MPE}.
Moreover, a direct optimization on the whole unlabeled data is generally inefficient, especially for implicit CF, where an efficient training approach supporting large-scale data is necessary~\cite{nnPU,sansone2018efficient}.
In our proposed solution, above issues are avoided by efficiently sampling negative instances from the unlabeled data and, motivated by the idea of PU learning, we carefully distinguish those true negative instances from others.

\textbf{Learning from Noisy Labels.}
By regarding unobserved instances as a combination of true negative labels and noisy labels, another choice is adapting the implicit CF into learning from noisy labels.
Typical learning approaches include curriculum learning~\cite{curriculum}, self-pace learning~\cite{selfpaced} and instance re-weighting~\cite{mentornet,L2Reweight,metaweight}.
The first two approaches prefer easier instances during training process so as to improve robustness, while these easy instances may be ineffective for learning a CF model.
Without prior information about the noisy labels, instance re-weighting approach learns the weight of each instance with bi-level optimization on training and validation data~\cite{mentornet,L2Reweight,metaweight}. However, the size of unlabeled data in implicit CF can approach to nearly a product of user count and item count, making above non-sampling approach become unaffordable in terms of learning efficiency. 
Therefore, this work focuses on negative sampling and aims to handle noisy labels correctly at the same time.

\subsection{Specific Negative Sampling Approaches}
Negative sampling approaches have also been widely adopted in other domains of embedding learning for text, graph, etc.
Motivated by these works that tend to leverage a simple model for capturing negative sampling distribution, we design a memory-based model that simply maintains the promising candidates with large scores.
More importantly, we propose to robustify negative sampling by emphasizing high-variance samples, which is novel in both CF and other domains.
The following table and review on literatures discuss the differences between different approaches.

\begin{table}[ht]
	\small
	\centering
	\begin{tabular}{C{45px} C{70px} C{65px} C{70px} C{90px}}
		\toprule
		Domain                                                               & Text                                                                           & Graph                                                                & Knowledge Graph                                                                                   & Collaborative Filtering \\ \midrule
		Learning Objective        & semantic word relationships         & node proximities          & fact composed of head/tail entity and relation           & user preferences among items        \\ \midrule
		Vanilla Sampling Strategy           & frequency-based \cite{word2vec}                                                                                                   & degree-based \cite{deepwalk,line}                                                                                               & uniform~\cite{TransE}, bernoulli~\cite{TransH}                                       & uniform~\cite{BPR}, popularity-based~\cite{NNCF,NCE4CF}                                                                                      \\ \midrule
		Improving Sample Quality & GAN~\cite{ACE}  & self-paced learning, GAN~\cite{selfpacedNE}  &  score-based~\cite{NSCaching}, GAN~\cite{ACE,KBGAN}                     & score-based~\cite{AdaOverSam,DNS}, GAN~\cite{RNS,AdvIR}  \\ \midrule
		Leveraging Skewness in Distribution & favouring frequent words~\cite{word2vec}                                                                                                   &  favouring high-degree nodes~\cite{deepwalk,line} or positive-alike nodes~\cite{yang2020understanding} & favouring large-scored instances~\cite{NSCaching} & none \\ \midrule
		Handling False Negative   & none      & none     &  none & avoiding the hardest instances~\cite{ying2018graph}    \\ \bottomrule
	\end{tabular}
\end{table}

\textbf{Negative Sampling in Other Domains.}
Negative sampling approaches are widely used in many tasks like word embedding~\cite{word2vec}, 
graph embedding~\cite{survey_ge} and knowledge graph embedding~\cite{survey_kge}.
In terms of capturing the distribution of negative instances, these applications generally requires a rather simple model.
For example, Word2Vec~\cite{word2vec} sets the negative sampling distribution proportional to the 3/4 power of word frequency, which favours those frequent words. Later works on graph embedding~\cite{deepwalk,line} readily keep this skewed distribution by adapting it to the node degree.
Similarly in knowledge graph, it has been observed that negative instances with large scores are important but rare and focusing on this partial set makes the model much simpler~\cite{NSCaching}.
Another recent work on negative sampling of graph representation learning proposes that the negative sampling distribution should be positively but sub-linearly correlated to their positive sampling distribution~\cite{yang2020understanding}.
However, in terms of avoiding false negative instances, none of them have tackled this problem by designing a robust sampling approach.
Since the implicit CF is a different problem where the reliability of sampled negative instances is much harder to guarantee, we propose to reduce this risk by emphasizing high variance samples. Meanwhile, motivated by above examples in other domains, we also leverage a simple model to efficiently capture the distribution of negative instances which are of high-quality.

\section{Implementation Details}
\subsection{Running Environment}
The experiments are conducted on a single Linux server with AMD Ryzen Threadripper 2990WX@3.0GHz, 128G RAM and 4 NVIDIA GeForce RTX 2080TI-11GB. Our proposed SRNS is implemented in Tensorflow 1.14 and Python 3.7.

\subsection{Baselines}\label{appendix:baseline}
We compare the SRNS with following state-of-the-art approaches:
(1) Uniform~\cite{BPR}, which uniformly selects negative samples from the unlabeled data. 
(2) NNCF~\cite{NNCF}, which uses a negative sampling distribution proportional to the 3/4 power of item popularity. A hyper-parameter $s$ is the number of positive samples per item. $b$ is the number of negative samples per positive sample.
(3) AOBPR~\cite{AdaOverSam}, which improves uniform strategy by adaptively oversampling hard instances. A hyper-parameter $\lambda$ controls the skewness of distribution $\propto \exp(-\text{rk}(j|u)/\lambda)$.
(4) IRGAN~\cite{IRGAN}, which uses an adversarial sampler by conducting a minimax game between the recommender and the sampler. A hyper-parameter $\tau$ is the temperature in sampling distribution~(Eq.~(10) in \cite{IRGAN}).
(5) RNS-AS~\cite{RNS}, which leverages adversarial sampling to generate hard negative samples. A hyper-parameter $N_s$ is size of candidate set for sampling and $\tau$ is the temperature.
(6) AdvIR~\cite{AdvIR}, which exploits both adversarial sampling and training (\textit{i.e.}, adding perturbation) to generate better negative samples. $N_s$ and $\tau$ are defined similarly as above. $\epsilon$ controls the perturbation size.
(7) ENMF~\cite{ENMF}, as a baseline, we also compare with an non-sampling approach that regards all the unlabeled data as negative labels and carefully assigns instance weights. A hyper-parameter $c$ controls above weight for a negative instance.

\subsection{Detail of MLP based $r$}\label{appendix:mlp}
The MLP based scoring function $r(\textbf{p}_u, \textbf{q}_i, \bm{\beta})$~\cite{NCF} takes the concatenation of $\textbf{p}_u$ and $\textbf{q}_i$, \textit{i.e.}, $\textbf{z}_0=[\textbf{p}_{u};\textbf{q}_{i}] \in \mathbb{R}^{2F}$, as the input.
Then there are $H$ hidden layers, and the $l$th layer is defined as 
\begin{equation}
\textbf{z}_l  = \mathrm{sigmoid}(\textbf{W}_l \textbf{z}_{l-1} + \textbf{b}_l),
\end{equation}
where $\textbf{W}_l\in \mathbb{R}^{d_l \times d_{l-1}}$ and $\textbf{b}_l\in \mathbb{R}^{d_l}$ denote the weight matrix and bias vector in this layer. Specifically, $d_0=2F$ and we set $d_l=\frac{1}{2}d_{l-1}$.
The last layer outputs the prediction score $r_{ui}$, defined as
\begin{equation}
    r_{ui}=\textbf{W}_{H+1}^{\top} \textbf{z}_H + \textbf{b}_{H+1},
\end{equation}
where $\textbf{W}_{H+1} \in \mathbb{R}^{d_H}$ and $\textbf{b}_{H+1}\in \mathbb{R}$.
The learnable parameters $\bm{\beta}$ in this MLP based $r$ are $\{\textbf{W}_i,\textbf{b}_i\} (i=1,...,H+1)$.

\subsection{Hyper-parameter Tuning}\label{appendix:para-tune}
Our SRNS's hyper-parameters can be divided into three parts: (1) sampling related part, including memory size $S_1$, expansion size $S_2$, temperature $\tau$, variance-based criterion weight $\alpha$, warm-start epoch number $T_0$.
(2) $r$ related part, including embedding dimension $F$ and hidden layer number $H$. (3) optimization related part, including learning rate $lr$ and L2 regularization $reg$.

In synthetic noise experiments, since we do not explicitly split a validation set on synthetic data, we draw two different train/test splits. The hyper-parameters are searched in the first round and afterwards are kept constant in another round. Note that the false negative instances~($\mathcal{F}$) in there two rounds are also independent with each other, as they are simulated by random sampling from the corresponding test set.
We run each synthetic data experiment 400 epochs without early stopping and repeat five times.
The scoring function $r$ is GMF~\cite{NCF}. The memory size $S_1/S_2$ are fixed as 20/20. The temperature $\tau$ is 1. Adam optimizer with $\beta_1=0.9$, $\beta_2=0.999$ is used and the mini-batch size is set to 1024. The lazy-update epoch number $E=1$. 
The rest hyper-parameters are tuned according to average NDCG@3 in the last 50 training epochs. 
Specifically, 
first we use grid search to find the best group of non-sampling related hyper-parameters, \textit{i.e.}, $(F,lr,reg)$, using the vanilla Uniform method~\cite{BPR} as the negative sampling strategy. Then we fix $(F,lr,reg)$ and search the rest sampling related hyper-parameters, \textit{i.e.}, $(\alpha, T_0)$, under different settings of noisy supervision~($\sigma$). See Table~\ref{tab:tune:syn} for detailed information.

\begin{table}[h]
    \small
    \setlength\tabcolsep{1pt}
    \centering
    \caption{SRNS's hyper-parameter exploration in synthetic noise experiments~(Section 4.2)}
    \begin{tabular}{C{80px} C{150px} C{70px}  C{70px} }
    \toprule
    Hyper-parameter & Tuning Range & Opt.~(Ecom-toy) & Opt.~(ML100k) \\ 
    \midrule
    $lr$&$[5, 10, 50]\times 10^{-4}$&0.001&0.001  \\
    $reg$&$[0, 1, 10]\times 10^{-3}$&0.0&0.001 \\
    $F$&$[8, 16, 32]$&32&8 \\
    $\alpha$&$[5.0,10.0,20.0,50.0]$&- &- \\
    $T_0$&$[50,100]$&- &- \\
    \bottomrule
    \end{tabular}
    \label{tab:tune:syn}
\end{table}

In real data experiments, we conduct the standard procedure to split train/validation/test set.
We run 400 epochs and terminate training if validation performance does not improve for 100 epochs, which has also been repeated five times. Both GMF and MLP~(defined in Appendix~\ref{appendix:mlp}) are tested. Adam optimizer with $\beta_1=0.9$, $\beta_2=0.999$ is used and the mini-batch size is set to 1024. The lazy-update epoch number $E=1$. The embedding dimension $F$ is set as 32~(ML-1m), 16~(Pinterest) and 8~(Ecom), respectively. We further show similar results with different $F\in\{8,16,32,64\}$ in Appendix~\ref{appendix:real}.
The rest hyper-parameters are tuned according to the best NDCG@1 on the validation set. 
Specifically,
first we use grid search to find the best group of non-sampling related hyper-parameters, \textit{i.e.}, $(lr,reg)$, using the vanilla Uniform method~\cite{BPR} as the negative sampling strategy~(For MLP based $r$ we also search $H$). 
Then we fix them and search the rest sampling related hyper-parameters, \textit{i.e.}, $(\tau,\alpha, T_0, S_1, S_2/S_1)$.
To ease the tuning process, we first fix $\alpha$ and $T_0$ as 0~(difficulty-only sampling), then search the best $(\tau, S_1, S_2/S_1)$.
After that we fix them and search the best group of $(\alpha,T_0)$~(variance-based sampling). Also, we repeat above step by changing memory size $S_1$ to its next or previous value. For example, if current best $S_1$ is 16, we further test 8 and 32.
Note that we do not search sampling related hyper-parameters when using MLP based $r$, by directly using those for GMF.
See Table~\ref{tab:tune:real} for detailed information.

\begin{table}[h]
    \small
    \setlength\tabcolsep{0.5pt}
    \centering
    \caption{SRNS's hyper-parameter exploration in real data experiments~(Section 4.3)}
    \begin{tabular}{C{30px} C{30px} C{150px} C{60px}  C{60px}  C{60px}}
    \toprule
    & Para. & Tuning Range & Opt.~(Ecom) & Opt.~(ML1m) & Opt.~(Pinterest) \\ 
    \midrule
    \multirowcell{7}{SRNS\\GMF}&$lr$&$[5, 10, 50, 100]\times 10^{-4}$&0.001&0.001&0.001  \\
	&$reg$&$[0, 1, 10, 100]\times 10^{-4}$&0.001&0.01&0.0 \\
	&$\tau$&$[0.5, 1.0, 2.0, 10.0]$&2.0&10.0&10.0  \\
	&$\alpha$&$[0.1, 1.0, 2.0, 5.0, 10.0, 20.0, 50.0]$&0.0&5.0&5.0 \\
	&$T_0$&$[25, 50, 100]$&25&50&50 \\
	&$S_1$&$[2, 4, 8, 16, 32]$&8&8&16 \\
	&$S_2/S_1$&$[1, 2, 4, 8]$&2&8&4 \\
	\midrule
	\multirowcell{3}{SRNS\\MLP}&$lr$&$[5, 10, 50]\times 10^{-4}$&0.001&0.001&-  \\
    &$reg$&$[0, 1, 10, 100]\times 10^{-4}$&0.001&0.01&- \\
    &$H$&$[0,1,2,3]$&3&3&- \\
    \bottomrule
    \end{tabular}
    \label{tab:tune:real}
\end{table}

As for the baselines listed in Appendix~\ref{appendix:baseline}, except Uniform~\cite{BPR} that have been tuned as above, others have also been carefully tuned according to their validation NDCG@1. For IRGAN, RNS-AS and AdvIR using GAN-based structure, we use a pretrained model~(\textit{i.e.}, trained under Uniform) to initialize.
See Table~\ref{tab:tune:real-baseline} for detailed information.

\begin{table}[t]
	\small
	\setlength\tabcolsep{1pt}
	\centering
	\caption{Baselines' hyper-parameter exploration in real data experiments~(Section 4.3)}
	\begin{tabular}{C{40px} C{30px} C{150px} C{50px}  C{50px} C{60px} }
	\toprule
	Method & Para. & Tuning Range & Opt.~(Ecom) & Opt.~(ML1m) & Opt.~(Pinterest) \\ 
	\midrule
	\multirowcell{2}{Uniform} &$lr$&$[5, 10, 50, 100]\times 10^{-4}$&0.001&0.001&0.001  \\
	&$reg$&$[0, 1, 10, 100]\times 10^{-4}$&0.001&0.01&0.0 \\
	\midrule
	\multirowcell{4}{NNCF} &$lr$&$[5, 10, 50, 100]\times 10^{-4}$&0.001&0.001&0.001  \\
	&$reg$&$[0, 1, 10, 100]\times 10^{-4}$&0.0&0.0&0.0 \\
	&$b$&$[32, 64, 128, 256, 512, 1024, 2048]$&32&2048&2048 \\
	&$s$&$[1, 2, 4]$&2&2&2 \\
	\midrule
	\multirowcell{3}{ENMF} &$lr$&$[5, 10, 50, 100]\times 10^{-4}$&0.01&0.01&0.005  \\
	&$reg$&$[0, 1, 10, 100]\times 10^{-4}$&0.001&0.0001&0.0 \\
	&$c$&$[0.01, 0.03, 0.05, 0.07, 0.1, 0.3, 0.5, 0.7]$&0.1&0.3&0.01  \\
	\midrule
	\multirowcell{3}{AOBPR} &$lr$&$[5, 10, 50, 100]\times 10^{-4}$&0.0005&0.0005&0.0005  \\
	&$reg$&$[0, 1, 10, 100]\times 10^{-4}$&0.001&0.01&0.0 \\
	&$\lambda$&$[5, 10, 20, 50, 100, 200, 500, 1000, 2000]$&10&1000&2000  \\
	\midrule
	\multirowcell{3}{IRGAN} &$lr$&$[5, 10, 50, 100]\times 10^{-4}$&0.0005&0.0005&0.0005  \\
	&$reg$&$[0, 1, 10, 100]\times 10^{-4}$&0.001&0.001&0.001 \\
	&$\tau$&$[0.5, 1.0, 2.0]$&2.0&1.0&1.0  \\
	\midrule
	\multirowcell{4}{RNS-AS} &$lr$&$[5, 10, 50, 100]\times 10^{-4}$&0.001&0.0005&0.0005  \\
	&$reg$&$[0, 1, 10, 100]\times 10^{-4}$&0.0&0.001&0.01 \\
	&$\tau$&$[0.5, 1.0, 2.0, 10.0]$&1.0&0.5&0.5  \\
	&$N_s$&$[10, 20, 30, 40]$&10&10&10 \\
	\midrule
	\multirowcell{5}{AdvIR} &$lr$&$[5, 10, 50, 100]\times 10^{-4}$&0.0005&0.0005&0.0005  \\
	&$reg$&$[0, 1, 10, 100]\times 10^{-4}$&0.0&0.0001&0.001 \\
	&$\epsilon$&$[1, 10, 100]\times 10^{-2}$&0.01&0.01&0.01  \\
	&$\tau$&$[0.5, 1.0, 2.0, 10.0]$&1.0&1.0&1.0 \\
	&$N_s$&$[10, 20, 30, 40]$&10&10&10 \\ 
	\bottomrule
	\end{tabular}
	\label{tab:tune:real-baseline}
\end{table}

\subsection{Evaluation Metrics}
As defined in Section~4.1, our used metrics, \textit{i.e.}, Recall and NDCG, can provide a comprehensive evaluation of model performance. The former measures whether the ground truth item is presented on the ranked list, while the latter measures the performance at a finer granularity by accounting for the position of hit.
The two datasets~(ML-100k and Ecom-toy) used in synthetic noise experiments are rather small, with the item count $|\mathcal{I}|$ between 1000$\sim$2000, while the rest three in real data experiments are much larger, with the highest value as 59290~(Ecom). 
Thus in real data experiments, we follow a common strategy~\cite{NCF,koren2008factorization} to fix the list length $|\mathcal{S}_u|$ as 100, by randomly sampling $100-|\mathcal{G}_u|$ non-interacted items, because ranking the whole item set for each user is too time-consuming during evaluation. 
When reporting NDCG@k and Recall@k, we choose a rather small value of truncated length $k\in\{1,3\}$, because of following two reasons:
(1) In real applications of implicit CF like recommender systems, users tend to browse the items at first few positions of a list, making the accuracy of rest recommended items less important.
(2) In real data experiments we fix the length of a ranked list as 100, thus choosing a large $k$ may make this task too easy.

\subsection{Variance Computation}
To calculate the prediction variance $\mathrm{std}[P_{\text{pos}}(k|u,i)]$~(Eq.~(4)) of each candidate instance $(u,k)$ stored in the memory $\mathcal{M}_u$, we directly use the prediction results from previous iterations, without any extra forward or backward passes in the $r$.
In our implementation, we consider the prediction probability in the latest few epochs, which is due to following two reasons:
(1) prediction history near the beginning period of training process is not stable for all kinds of negative instances, and thus can be excluded from the computation.
(2) this implementation makes the overhead constant~($O(1)$) for each sampling operation.
In our experiments, we determine the uncertainty only based on the latest 5 epochs. Specifically, at $t$th training epoch,
\begin{equation}
\begin{split}
\mathrm{std}[P_{\text{pos}}(k|u,i)] &= \sqrt{
\nicefrac{
\sum_{s=t-5}^{t-1} 
\left[
\left[ P_{\text{pos}}(k|u,i) \right]_s - \text{Mean}\left[ P_{\text{pos}}(k|u,i) \right]
\right]^2} {5}
}, \\
\text{Mean}[ P_{\text{pos}}(k|u,i) ] &= \nicefrac{
\sum_{s=t-5}^{t-1} 
\left[ P_{\text{pos}}(k|u,i) \right]_s} {5}.
\end{split}
\end{equation}

In real data experiments where the datasets are much larger, it is time-consuming to compute the prediction probability~($P_{\text{pos}}$) for all user-item pairs~($|\mathcal{U}|\cdot |\mathcal{I}|$)  at each epoch.
Thus we prune the item space for each user's memory update process, so as to avoid logging $P_{\text{pos}}$ for all items.
Specifically, for $u$ at $t$th training epoch, the newly extended candidates in $\bar{\mathcal{M}}_u$ can only be randomly sampled from 
an item set, denoted as $var\_set_u$, which has already been generated at $(t-5)$th epoch. 
At the mean time, for $v \in var\_set_u$, we log $P_{\text{pos}}(v|u,i)$ values at the subsequent 5 epochs.
Therefore, among $u$'s memory $\mathcal{M}_u$, besides the original items that have been maintained from previous epochs, the newly added items also have the $P_{\text{pos}}$ history in the latest 5 epochs, which supports the variance computation above.
Note that $var\_set_u$ is also generated by random sampling from $u$'s non-interacted items, and its size is larger than memory size $S_1$, but much smaller than item count $|\mathcal{I}|$. 
We fix $|var\_set_u|$ as 3000~(ML-1m) and 600~(Pinterest, Ecom), respectively.

\section{Experiment Details}
\subsection{Dataset Description}

We choose following four raw datasets and build five datasets for performance evaluation.
\begin{itemize}
\item \textbf{Movielens~(ML)-100k}\footnote{https://grouplens.org/datasets/movielens/100k}. This is a widely used movie-rating dataset containing 100,000 ratings on movies from 1 to 5. We follow the common preprocessing to convert it into implicit feedback data, regarding those high-rated records~($4\sim5$) as positive labels~\cite{AdvIR,IRGAN}.

\item \textbf{Movielens~(ML)-1m}\footnote{https://grouplens.org/datasets/movielens/1m}. Similarly to ML-100k, this large dataset contains 1,000,000 ratings. After similar converting procedure, we filter out users with less than 5 records.

\item \textbf{Pinterest}\footnote{https://pinterest.com}. This implicit feedback dataset is constructed by \cite{geng2015learning} for a task of image recommendation, and has been used for evaluating the implicit CF task~\cite{NCF}.

\item \textbf{Ecommerce~(Ecom)}. This implicit feedback dataset is a subset of users' item-click records in a real-word E-commerce website between 2017/06 and 2017/07. For data preprocessing, we filter out users/items with less than 4 records, so as to overcome the problem of high sparsity. After that, we further obtain a toy dataset, denoted as \textbf{Ecom-toy}, by retaining top 1,000 users and 2,000 items sorted by number of records.
\end{itemize}

\subsection{Details of Figure 1}\label{appendix:motivation}
The experiment is conducted on ML-100k dataset, using the same train/test split as synthetic noise experiments. We use GMF as the $r$ and Uniform~\cite{BPR} as the negative sampling strategy.
By flipping labels of groundtruth records in the test set, we are able to obtain a set of false negative instances~(FN) that are in fact positive labeled but unobserved during the negative sampling process.
Besides uniformly sampling negative instances~(UN) to update the model, we simultaneously obtain a series of hard negative instances~(HN) with different difficulty $D$.
In following analysis, we adopt a simple yet effective strategy to control $D$ of a obtained HN: 1) uniformly sample $D$ candidates from $\{(u,j)|j\notin \mathcal{R}_u\}$; 2) select the negative instance with the highest value of $r_{uj}$. When $D$ gets higher, HN becomes much harder. UN is the same as HN with  $D=1$.

As in Figure~1(a), we have a closer look at the negative instances' distribution in terms of their positive-label probabilities $P_{\text{pos}}$ that are proportional to the prediction scores. This is motivated by \cite{NSCaching} that has observed a skewed distribution of negative instances when learning knowledge graph embeddings. 
Specifically, (a) is the distribution of negative instances $\{(u,j)| u\in \mathcal{U}, j\notin \mathcal{R}_u\}$ at 5 timestamps. We measure the \textit{complementary cumulative distribution function}~(CCDF) $F(x)=P(P_{\text{pos}} \geq x)$ to show the proportion of negative instances that satisfy $P_{\text{pos}} \geq x$.
Since hard negative instances generally have large $P_{\text{pos}}$, we compare them with those false negative instances \textit{w.r.t.} $P_{\text{pos}}$~(Figure~1(b)). We use the median value~($p50$) to represent each set. 
Then in Figure~1(c), we further analyze the possibility of using $P_{\text{pos}}$ to discriminate above two sets of negative instances. Specifically, under different hard negative sampling strategies, we calculate the \textit{label error ratio} in each mini-batch, \textit{i.e.}, $LER=\nicefrac{(\#\ of\ false\ negative\ samples)}{(\#\ of\ all\ selected\ negative\ samples)}$.
Unlike others, false negative instances follow the similar distribution as those positive instances in training data. Thus the model can ideally become more and more confident about predicting them as positive instances, and the corresponding variance of $P_{\text{pos}}$ is low.
Finally, to validate this, we compare $P_{\text{pos}}$'s variance between different types of negative instances in Figure~1(d). The normalized variance is measured by the ratio between standard deviation and mean value. 

\begin{figure*}[b]
\centering
\subfigure[$\sigma=0$, ML-100k]{\includegraphics[width=.25\textwidth]{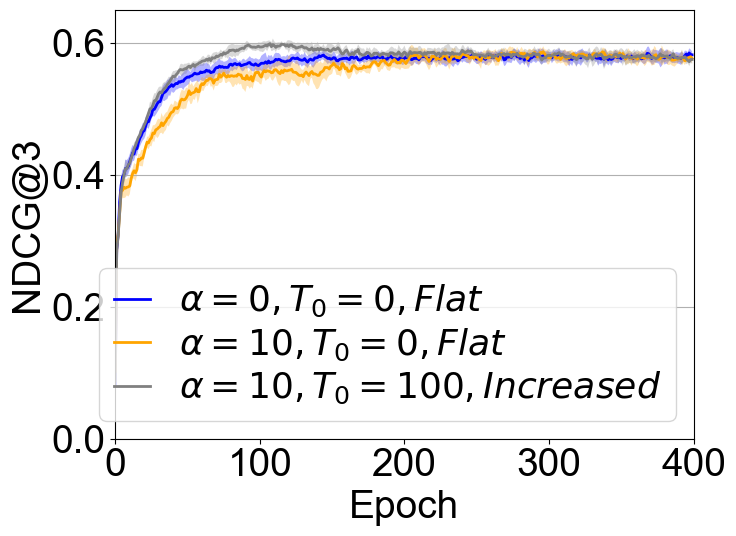}}\hfill
\subfigure[$\sigma=0.2$, ML-100k]{\includegraphics[width=.25\textwidth]{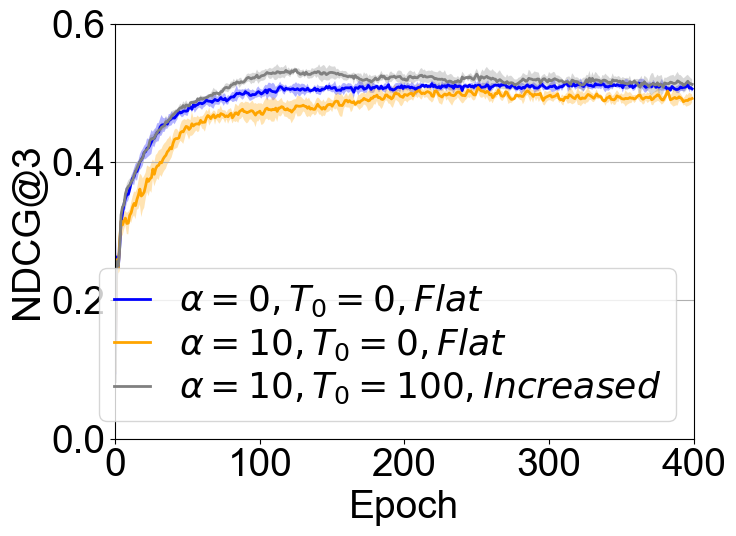}}\hfill
\subfigure[$\sigma=0.6$, ML-100k]{\includegraphics[width=.25\textwidth]{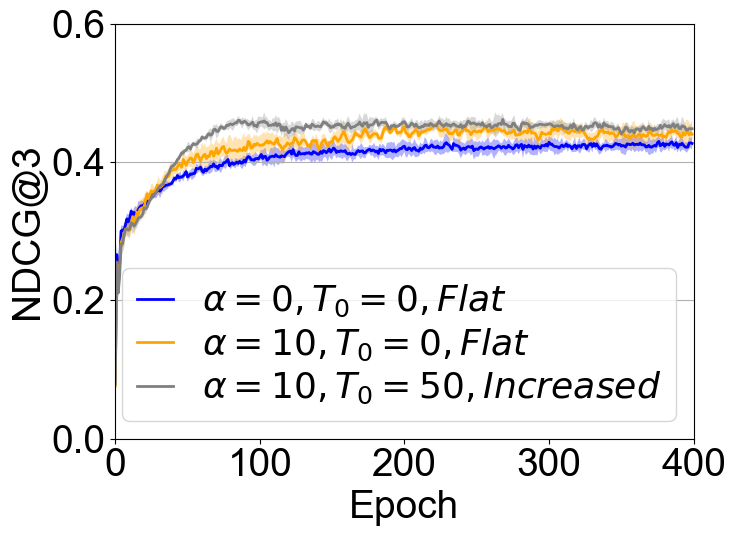}}\hfill
\subfigure[$\sigma=0.8$, ML-100k]{\includegraphics[width=.25\textwidth]{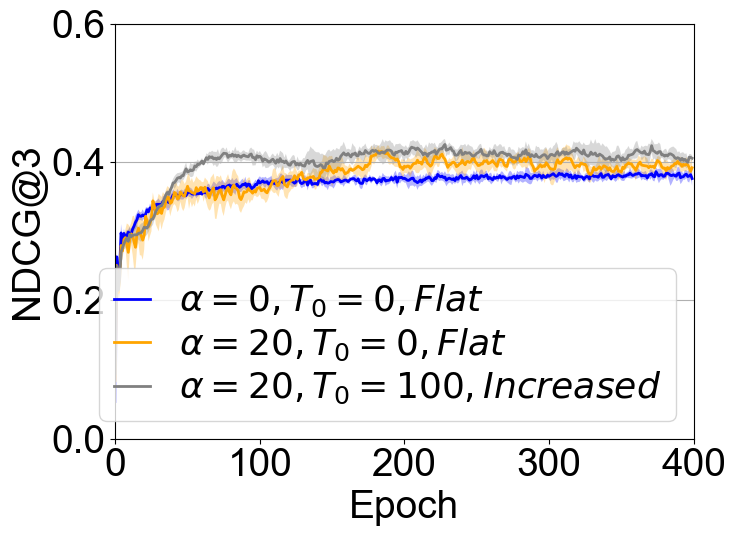}}
\vfill
\subfigure[$\sigma=0$, Ecom-toy]{\includegraphics[width=.25\textwidth]{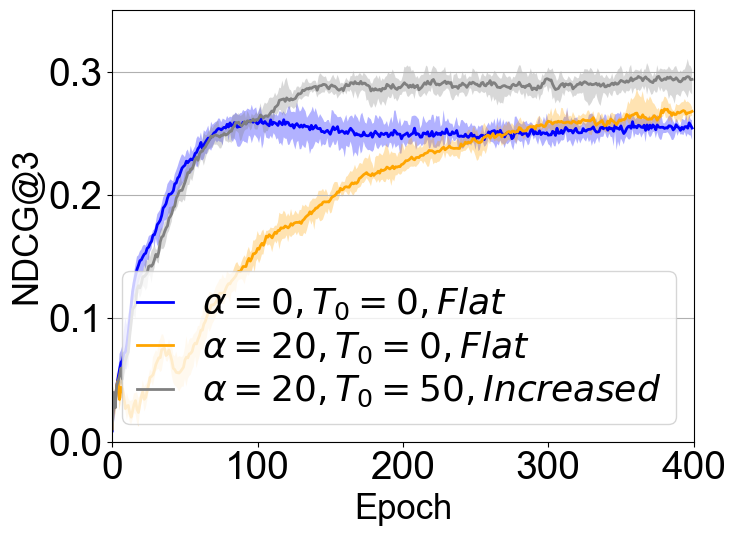}}
\subfigure[$\sigma=0.3$, Ecom-toy]{\includegraphics[width=.25\textwidth]{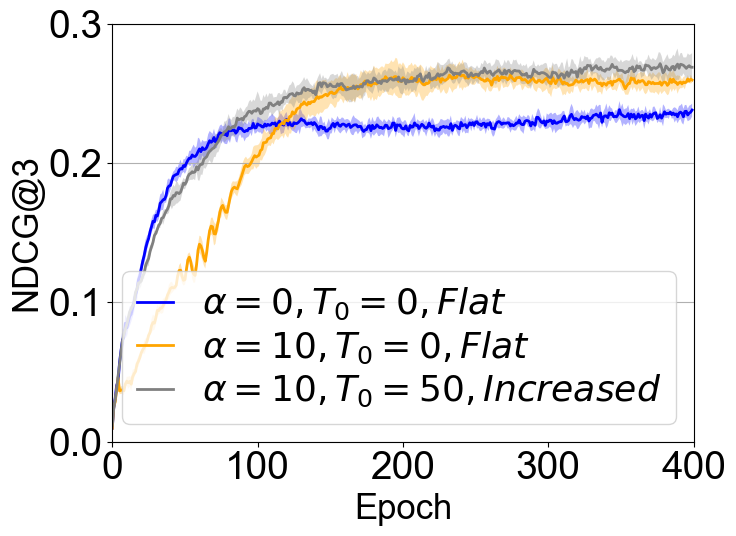}}\hfill
\subfigure[$\sigma=0.7$, Ecom-toy]{\includegraphics[width=.25\textwidth]{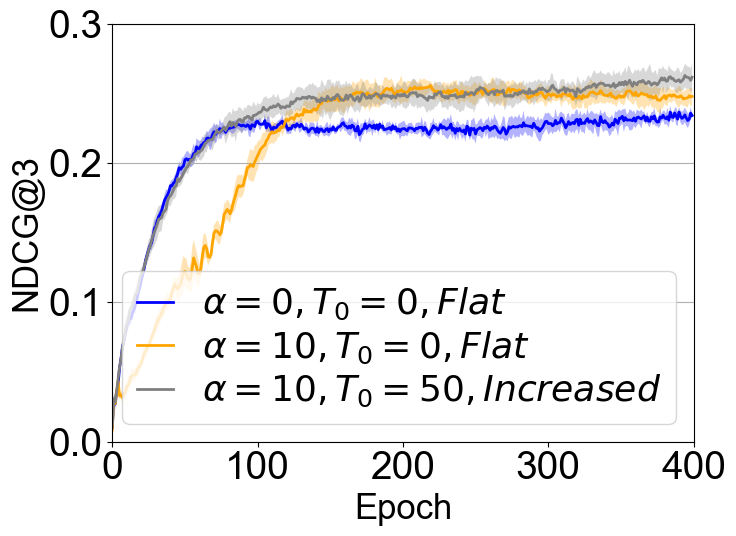}}\hfill
\subfigure[$\sigma=1$, Ecom-toy]{\includegraphics[width=.25\textwidth]{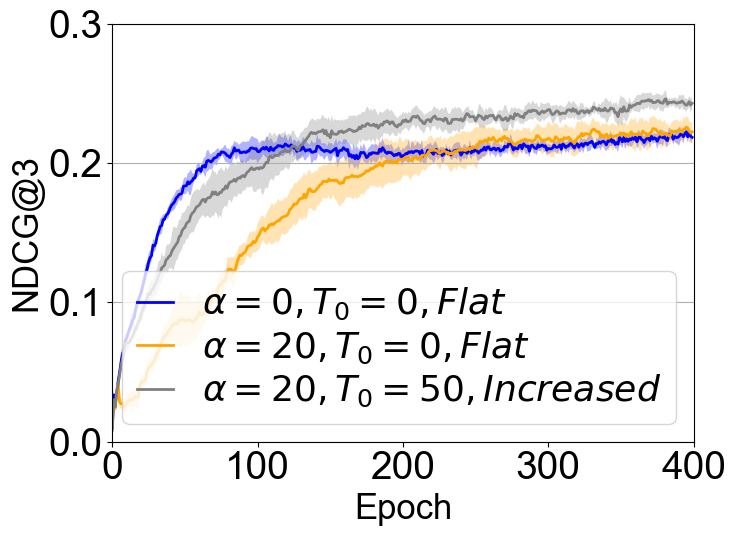}}
\caption{Detailed results of Figure 3: Test NDCG vs. number of epochs on two datasets, with the error bar for STD highlighted as a shade.}	
\label{fig_synthetic3}
\end{figure*}
\subsection{Synthetic Noise Experiments}\label{appendix:synthetic}
To control the impact of false negative instances on the sampling process, 
we manually inject noisy labels by slightly modifying each user's memory $\mathcal{M}$ that stores $S_1$ candidate negative instances. 
Specifically, for user $u$, there is always an instance in $\mathcal{M}_u$ that is randomly sampled from $u$'s false negative set $\mathcal{F}_u$, and this instance is also dynamically updated together with $\mathcal{M}_u$. As for the rest $S_1-1$  candidates in $\mathcal{M}_u$, they cannot be selected from $\mathcal{F}_u$.
To control the noise ratio, we vary the size of false negative set by randomly sampling $\sigma\times100$~(\%) from $\mathcal{F}_u$~($\sigma\in[0,1]$). Note that $\sigma=0$ indicates an ``ideal'' case where $\mathcal{M}_u$ is not influenced by $\mathcal{F}_u$.
In these experiments, we fix the memory size $S_1$ as 20.

Note that in the ``ideal'' case with no explicit noise, 
SRNS still largely outperforms in Ecom-toy dataset, 
which is also reasonable given the fact that $\mathcal{F}$ cannot ideally cover all the false negative instances hidden in unlabeled data.

In each figure of Figure 5, 
the blue curve represents the result of difficulty-only sampling strategy, 
while the grey curve and orange curve both represent those of the SRNS, 
with the difference on whether to linearly increase weight $\alpha_t$ during training process.
It can be clearly observed that the 
``warm-start'' setting of $\alpha_t$ performs better than a fixed-value setting, 
as the former better leverages prediction variance after false negative instances become stable.  
More detailed investigation on different settings of $\alpha_t$ are shown in following two tables.

\begin{table}[h]
    \small
    \setlength\tabcolsep{1pt}
    \centering
    \caption{Detailed investigation of ``warm-start'' on ML-100k, $\sigma=1.0$~(Figure 3(a)).}
    \begin{tabular}{C{60px} C{60px} C{60px}  C{60px} C{60px} C{60px}}
    \toprule
     & $T_0$/$\alpha$ & 5 & 10 & 20&50 \\  \midrule

    Flat &0   &0.3703$\pm$0.0033&   0.3811$\pm$0.0048&  0.3876$\pm$0.0054&  0.4004$\pm$0.0112 \\ \midrule
    \multirowcell{2}{Increased} & 50 &0.3734$\pm$0.0045&0.3931$\pm$0.0097&0.3924$\pm$0.0050&0.3965$\pm$0.0099\\ 
    & 100 &0.3725$\pm$0.0111&   0.3850$\pm$0.0075&  \textbf{0.4062$\pm$0.0073}& 0.3844$\pm$0.0078 \\ \midrule
    \multirowcell{2}{Decreased} &50 &0.3631$\pm$0.0066& 0.3677$\pm$0.0049&  0.3700$\pm$0.0064&  0.3623$\pm$0.0108\\ 
     & 100  &0.3620$\pm$0.0063& 0.3650$\pm$0.0039&  0.3710$\pm$0.0062&  0.3719$\pm$0.0055\\ 
    \bottomrule
    \end{tabular}
\end{table}

\begin{table}[h]
    \small
    \setlength\tabcolsep{1pt}
    \centering
    \caption{Detailed investigation of ``warm-start'' on Ecom-toy, $\sigma=0.5$~(Figure 3(b).}
    \begin{tabular}{C{60px} C{60px} C{60px}  C{60px} C{60px} C{60px}}
    \toprule
     & $T_0$/$\alpha$ & 5 & 10 & 20&50 \\  \midrule

    Flat &0   &0.2449$\pm$0.0052&   0.2557$\pm$0.010&   0.2525$\pm$0.0063&  0.2343$\pm$0.0019 \\ \midrule
    \multirowcell{2}{Increased} & 50 &0.2574$\pm$0.0051&0.2702$\pm$0.0048&0.2515$\pm$0.0053&0.2329$\pm$0.0090\\ 
    & 100 &0.2464$\pm$0.0051&   \textbf{0.2581$\pm$0.0072}& 0.2636$\pm$0.0091&  0.2267$\pm$0.0092 \\ \midrule
    \multirowcell{2}{Decreased} &50 &0.2037$\pm$0.0064& 0.2351$\pm$0.0081&  0.2367$\pm$0.0091&  0.2365$\pm$0.0110\\ 
     & 100  &0.2120$\pm$0.0029& 0.2351$\pm$0.0062&  0.2348$\pm$0.0051&  0.2513$\pm$0.0053\\ 
    \bottomrule
    \end{tabular}
\end{table}

\subsection{Real Data Experiments}\label{appendix:real}
\begin{figure*}
\centering
\subfigure[$F$, N@1, Pinterest]{\includegraphics[width=.25\textwidth]{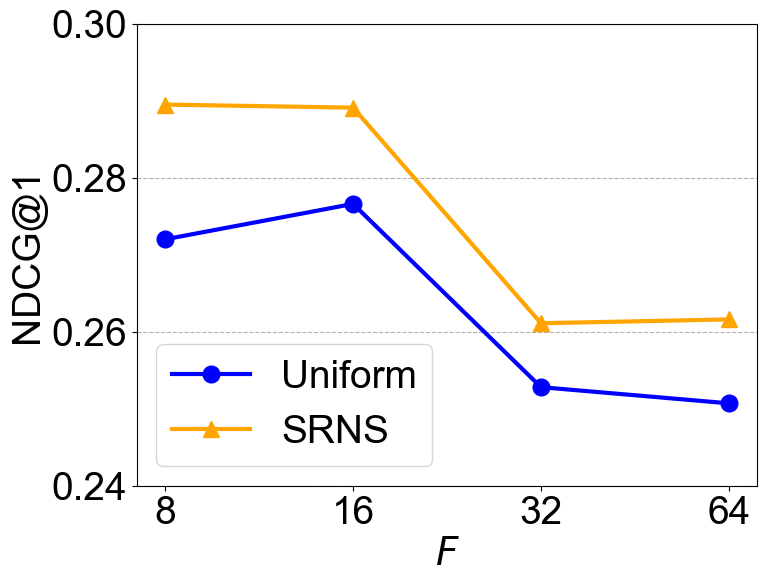}}
\subfigure[$F$, N@3, Pinterest]{\includegraphics[width=.25\textwidth]{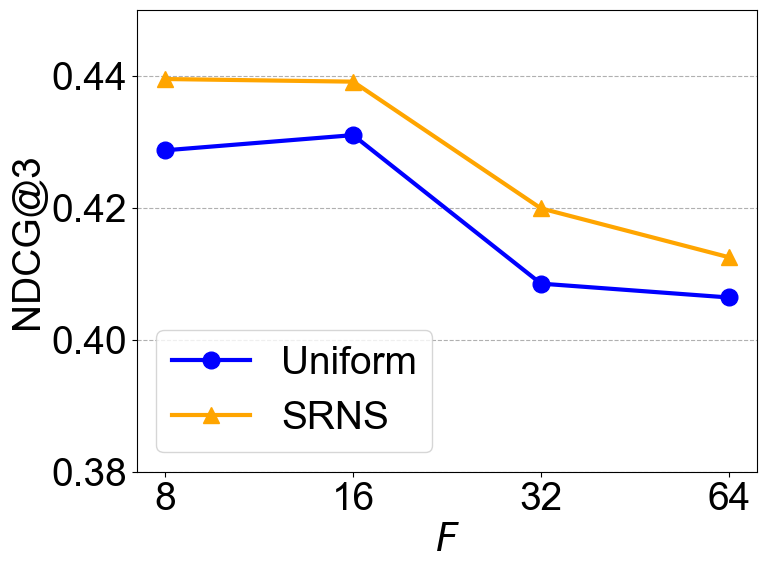}}
\subfigure[$F$, R@3, Pinterest]{\includegraphics[width=.25\textwidth]{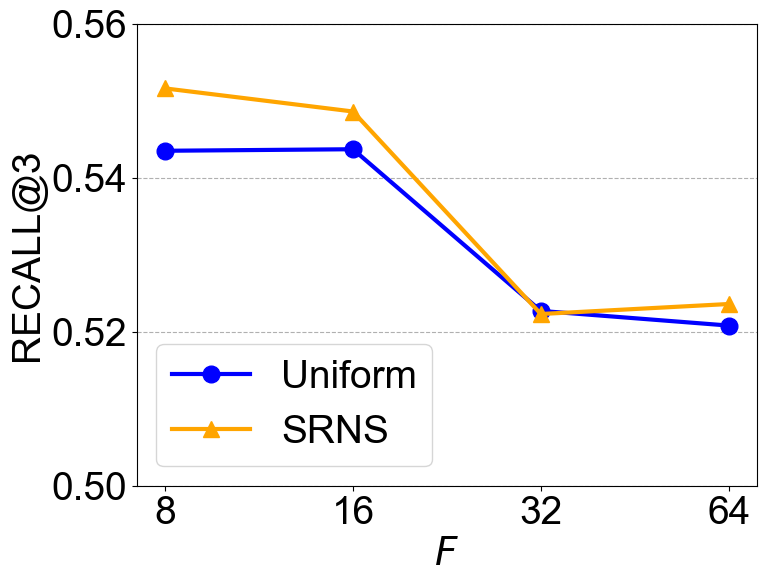}}
\vfill
\subfigure[$F$, N@1, Ecom]{\includegraphics[width=.25\textwidth]{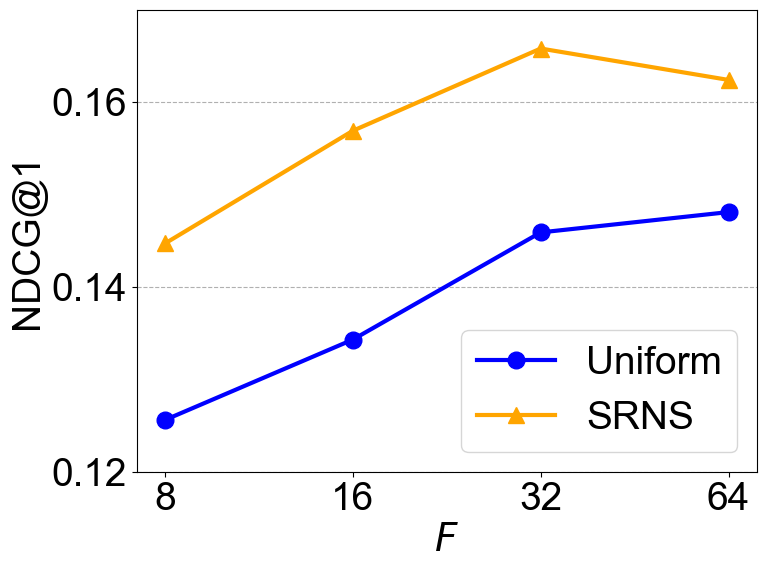}}
\subfigure[$F$, N@3, Ecom]{\includegraphics[width=.25\textwidth]{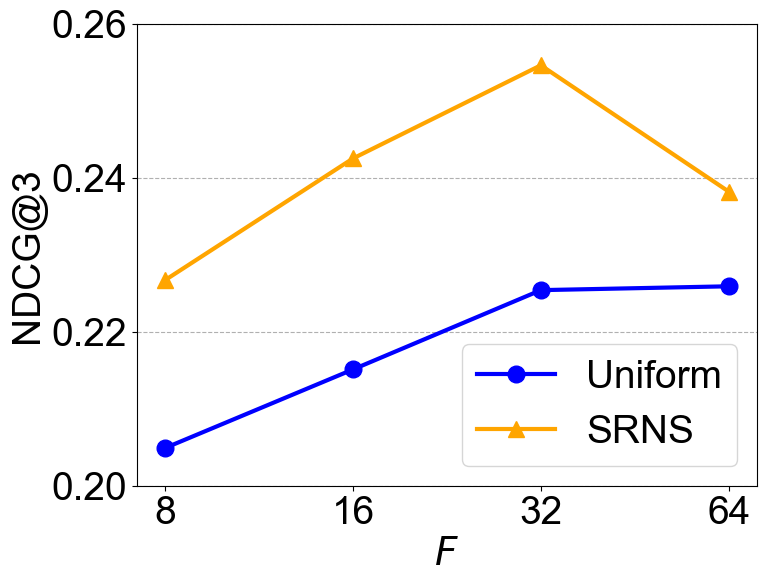}}
\subfigure[$F$, R@3, Ecom]{\includegraphics[width=.25\textwidth]{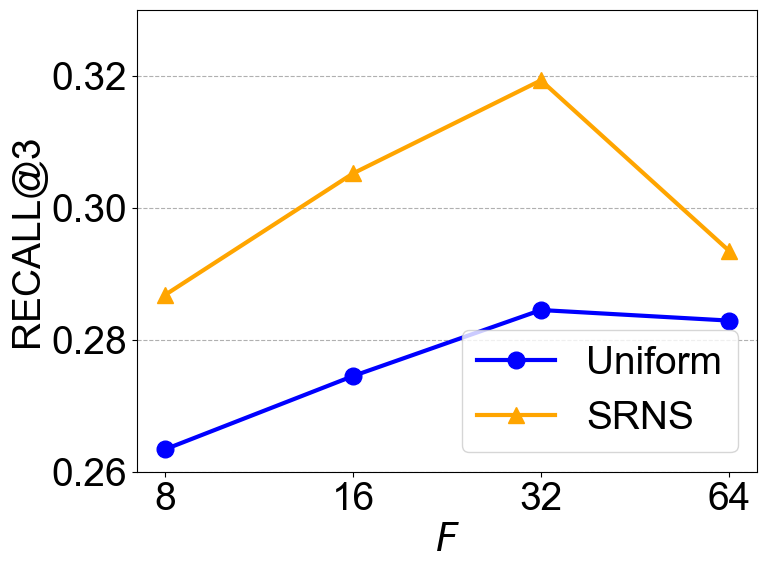}}
\caption{Varying embedding dimension $F$: Test NDCG/Recall of Uniform and SRNS approaches, using different embedding size $F$, on Pinterest and Ecom, respectively.}	
\label{fig_generality1}
\end{figure*}

Figure~\ref{fig_generality1} shows test NDCG of Uniform and SRNS approaches using different embedding size $F$. The scoring function $r$ is GMF. Again we can observe consistent improvement of SRNS over Uniform when $F\in\{8,16,32,64\}$. Although increasing $F$ should have improved performance, we observe instead that $F=16$ performs the best on Pinterest dataset, which conforms to a previous work~(Figure 4 in \cite{NCF}).

\begin{figure*}[t]
\centering
\subfigure[$S_1$, N@3, ML-1m]{\includegraphics[width=.25\textwidth]{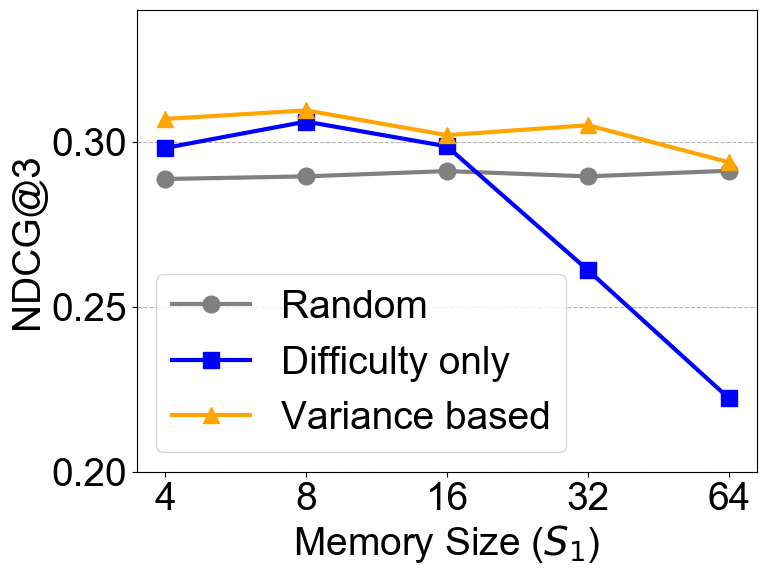}}
\subfigure[$S_1$, R@3, ML-1m]{\includegraphics[width=.25\textwidth]{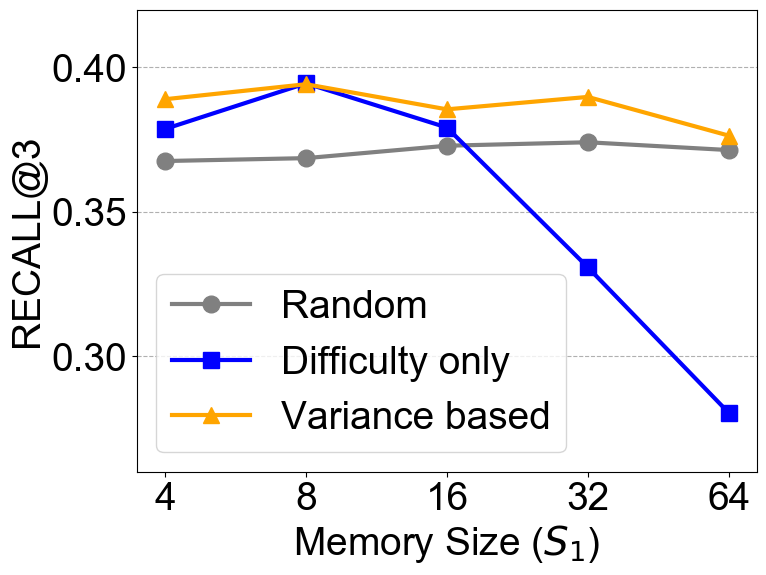}}
\subfigure[$S_1$, N@3, Pinterest]{\includegraphics[width=.25\textwidth]{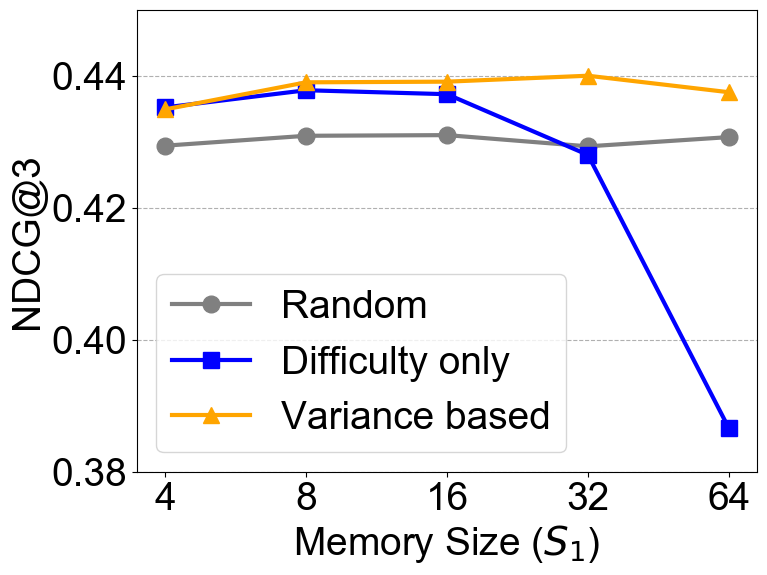}}

\vfill
\subfigure[$S_1$,R@3, Pinterest]{\includegraphics[width=.25\textwidth]{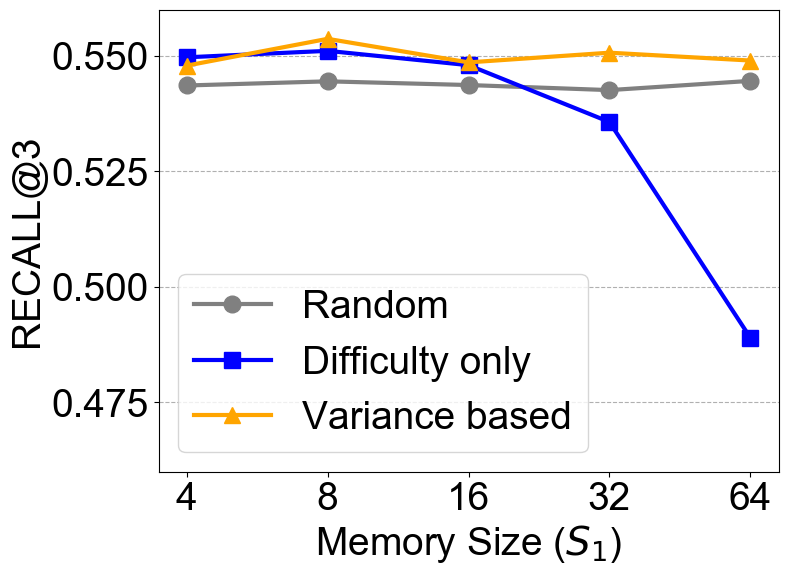}}
\subfigure[$S_1$, N@3, Ecom]{\includegraphics[width=.25\textwidth]{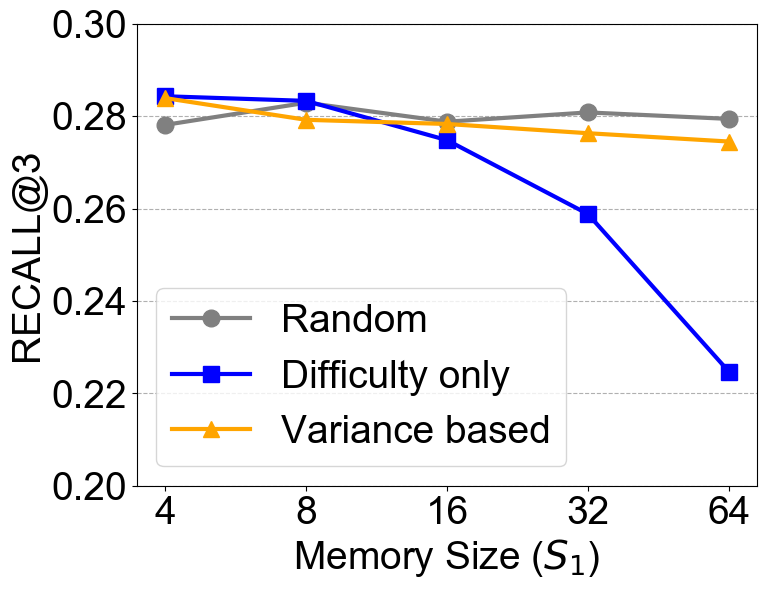}}
\subfigure[$S_1$, R@3, Ecom]{\includegraphics[width=.25\textwidth]{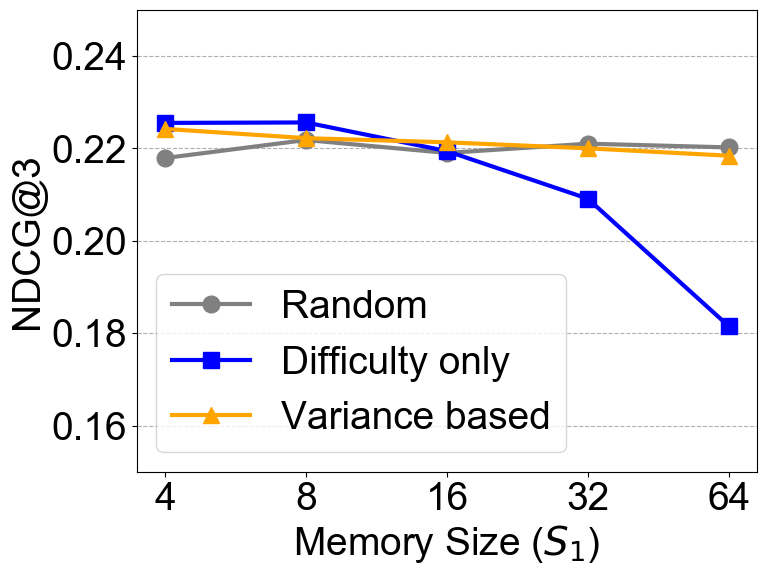}}
\caption{Detailed results of Figure 4(e) and (f): Test NDCG@3/Recall@3 vs. SRNS’s memory size $S_1$, using different sampling strategies on three datasets.}	
\label{fig_quality1}
\end{figure*}

Figure~\ref{fig_quality1} shows supplementary results, \textit{w.r.t.} NDCG@3 and Recall@3, of Figure~4(d)-(f), which are similar to those findings \textit{w.r.t.} NDCG@1.

\end{document}